\documentclass{article} % For LaTeX2e
\usepackage{iclr2027_conference,times}

\usepackage{amsmath,amsfonts,bm}

\def\eqref#1{equation~\ref{#1}}
\def\1{\bm{1}}

\DeclareMathAlphabet{\mathsfit}{\encodingdefault}{\sfdefault}{m}{sl}
\SetMathAlphabet{\mathsfit}{bold}{\encodingdefault}{\sfdefault}{bx}{n}

\usepackage{hyperref}
\usepackage{url}

\usepackage{graphicx}
\usepackage{multirow}
\usepackage{float}        % table[H] (정확히 이 자리에)
\usepackage{pgfplots}

\usepackage{amsthm}
\theoremstyle{plain}

\theoremstyle{definition}

\usepackage{cleveref}   % \cref — load after hyperref and after \newtheorem
\crefname{proposition}{Proposition}{Propositions}
\crefname{lemma}{Lemma}{Lemmas}
\crefname{corollary}{Corollary}{Corollaries}
\crefname{assumption}{Assumption}{Assumptions}
\crefname{definition}{Definition}{Definitions}

\title{Is Online Interaction Necessary for Recovery? A Minimalist Approach to Robust Planning via Perturbation}

\author{Bumgeun Park, Donghwan Lee\thanks{Corresponding authors}\\
Korea Advanced Institute of Science and Technology (KAIST)\\
\texttt{\{j4t123,donghwan\}@kaist.ac.kr}
}

\iclrfinalcopy % Uncomment for camera-ready version, but NOT for submission.
\begin{document}

\maketitle

\begin{abstract}
Behavior cloning (BC) is vulnerable to covariate shift during closed-loop execution, where small prediction or execution errors can drive the robot toward states poorly covered by the demonstration data. We focus on action-sequence planning, where a policy predicts a finite-horizon sequence of actions as a reference trajectory for robot execution. Existing approaches to covariate shift often rely on collecting additional corrective demonstrations, requiring further environment interaction and access to an expert or reference policy. We propose Perturbation-Augmented Recovery Supervision (PARS), a simple training approach for improving recovery using only existing demonstrations. PARS perturbs the robot's proprioceptive state and anchors only the later portion of the predicted trajectory to the original demonstration, leaving the earlier portion free to generate corrective motion and recover toward the demonstrated behavior within the planning horizon. Unlike conventional input-noise augmentation, which preserves the original supervision target over the entire prediction horizon, PARS explicitly provides trajectory-level supervision for recovery from perturbed states. PARS requires neither additional environment interaction nor expert queries and can be instantiated across different action-sequence policy classes with only minor modifications to their BC objectives. Experiments on 51 RLBench manipulation tasks with flow-based, transformer-based, and diffusion-based policies demonstrate that PARS improves robustness to covariate shift during closed-loop execution.
\end{abstract}

\section{Introduction}
Behavior cloning (BC) is a method for training a policy to reproduce desired behaviors from demonstrations, providing a direct way to acquire complex robotic skills without explicitly specifying reward functions \citep{osa2018algorithmic}. However, BC from (near-)expert demonstrations is susceptible to covariate shift during closed-loop execution, where small prediction or execution errors can drive the robot toward states poorly covered by the demonstration data and eventually lead to performance degradation. Several approaches have sought to mitigate this issue by improving robustness around the demonstrated distribution through corrective-label synthesis, distributionally robust objectives, stability-aware learning, and trajectory synthesis \citep{ke2024ccil, seo2024mitigating, agrawal2025markov, mehta2025stable, wang2026koopman}. Beyond single-step action prediction, predicting temporally extended action sequences over a finite horizon has become widely adopted in robotic BC and can further mitigate compounding errors \citep{lazzati2026does, zhao2023learning, chi2025diffusion}. More recent approaches additionally introduce explicit stabilization or filtering mechanisms to improve robustness against deviations during execution \citep{pmlr-v305-jiang25a, rosasco2025kdpe}.

Another line of work addresses distribution shift more directly by augmenting the training dataset with corrective demonstrations. \citet{ross2011reduction} iteratively rolls out the learned policy and queries an expert for corrective actions at visited states, thereby expanding the training distribution toward states encountered during deployment. Alternatively, \citet{laskey2017dart} inject perturbations into expert actions to induce likely error states and collect recovery behaviors. Subsequent interactive imitation learning methods reduce the burden of expert supervision through selective querying or intervention \citep{zhang2016query, kelly2019hg}. However, these approaches require additional environment interaction and continued access to an expert or reference policy, increasing the cost of applying them to real-world robotic systems.

In this paper, we consider planners that predict finite-horizon sequences of desired robot configurations as reference trajectories, a formulation widely used in robotic systems. We propose \textbf{P}erturbation-\textbf{A}ugmented \textbf{R}ecovery \textbf{S}upervision (PARS) to improve recovery from covariate shift without additional environment interaction. During training, PARS perturbs proprioceptive observations from existing demonstrations and anchors only the predicted trajectory suffix to the corresponding demonstration suffix. The preceding prefix receives no direct augmented target, leaving room for corrective motion before rejoining the demonstrated continuation. Unlike conventional input-noise augmentation, which retains the original target over the full horizon and primarily encourages invariance to perturbations \citep{ke2021grasping, ameperosa2025rocoda}, PARS specifies where the perturbed trajectory should return. The standard BC objective still trains the full action sequence on unperturbed demonstrations. PARS therefore requires no explicit recovery trajectories and can be incorporated into existing action-sequence policies with only minor modifications to their training objectives.

Our contributions are summarized as follows:
\begin{itemize}
    \item We propose PARS, a simple supervision scheme for improving recovery during closed-loop execution using only existing demonstrations, without additional environment interaction or expert queries during training.

    \item We provide a theoretical analysis of PARS, characterizing how suffix anchoring affects local sensitivity to proprioceptive perturbations across policy classes and deriving sufficient conditions for the free recovery prefix to remain close to demonstrated motion and satisfy joint-space speed and acceleration limits.

    \item Across 51 RLBench manipulation tasks and multiple action-sequence policy classes, we show that PARS improves mean closed-loop success and promotes recovery toward demonstrated behavior or mitigates further deviation.
\end{itemize}

\section{Related Works}
\subsection{Robustness in Single-Step Behavior Cloning}
BC commonly trains a policy to predict a single action conditioned on the current observation, making it susceptible to compounding errors under covariate shift. Several approaches have sought to improve the robustness of policies that predict individual actions. CCIL \citep{ke2024ccil} synthesizes corrective labels using learned local dynamics, DrilDICE \citep{seo2024mitigating} applies stationary-distribution correction, and other methods exploit structural or stability properties of demonstration trajectories \citep{agrawal2025markov, mehta2025stable}. More recently, KATS \citep{wang2026koopman} generates dynamically consistent trajectories in a learned Koopman latent space to augment the training distribution. As shown in Appendix~\ref{app:single-step}, these methods perform substantially worse than action-sequence policies in our setting; we therefore focus our main evaluation on action-sequence methods.

\subsection{Robustness through Action-Sequence Modeling}
An alternative direction is to move beyond single-step action prediction and generate temporally extended action sequences. Action-sequence modeling has become widely adopted in robotic BC, with recent analysis attributing its benefits to reduced compounding errors, increased non-Markovian expressivity, and implicit ensembling across temporal predictions \citep{lazzati2026does}. Action Chunking with Transformers (ACT) \citep{zhao2023learning} predicts action sequences using a Transformer-based conditional generative model, while Diffusion Policy (DP) \citep{chi2025diffusion} represents distributions over action sequences through a denoising diffusion process. More recent methods explicitly target robustness to deviations during execution. Streaming Flow Policy (SFP) \citep{pmlr-v305-jiang25a} represents action trajectories as flow trajectories and introduces a stabilizing flow field around demonstrated behaviors, whereas KDPE \citep{rosasco2025kdpe} generates multiple candidate action sequences from a diffusion policy and filters potentially harmful trajectories using kernel density estimation.

\subsection{Interactive Corrective Data Collection}
A direct way to address covariate shift is to expand the training distribution with states outside the original demonstration distribution and provide corrective supervision at those states. DAgger \citep{ross2011reduction} established this paradigm by iteratively rolling out the learned policy, querying an expert at visited states, and aggregating the newly labeled samples into the training dataset. Subsequent methods reduce the supervision burden through selective querying or intervention, as in SafeDAgger \citep{zhang2016query} and HG-DAgger \citep{kelly2019hg}. Rather than rolling out the learner, DART \citep{laskey2017dart} perturbs expert actions during demonstration collection to induce deviations from the expert trajectory and collect corresponding recovery behaviors. More recently, \citet{zhang2026action} analyzed an exploratory data-collection scheme that perturbs expert actions during data collection while retaining expert supervision at the resulting states, thereby collecting data around expert trajectories without rolling out the learned policy. Despite their different collection strategies, these approaches require additional environment interaction and expert or reference supervision to obtain corrective data. Our work instead investigates whether recovery supervision can be constructed directly from existing demonstrations without collecting additional interactive trajectories.

\section{Preliminaries}
\label{sec:preliminaries}
\paragraph{Problem Formulation.}
We consider a sequential decision-making problem in which the robot receives an observation $o_t=(o_t^{\mathrm{prop}},o_t^{\mathrm{env}})$, where $o_t^{\mathrm{prop}}$ contains the robot's proprioceptive state, including its current configuration and gripper state, and $o_t^{\mathrm{env}}$ describes the surrounding scene and other task-relevant information. In this work, we consider a robotic manipulator equipped with a gripper. Each action is represented as $a_t=(q_t^{\mathrm{des}},g_t)$, where $q_t^{\mathrm{des}}\in\mathbb{R}^{d_q}$ specifies the desired robot configuration, $d_q$ is the number of robot degrees of freedom, and $g_t\in\mathbb{R}$ specifies the gripper command. The desired configuration is subsequently tracked by a low-level controller, so that the policy specifies high-level reference commands rather than low-level control inputs. In our experiments, the policy parameterizes these configuration commands as joint displacements, which are accumulated from the configuration at the beginning of each action sequence to obtain the desired configurations.

\paragraph{Action-Sequence Planning.}
Rather than predicting a single action, we consider a planning policy that predicts an $H$-step sequence of future actions,
\begin{equation}
    \mathbf{a}_t
    =
    \left(a_t, a_{t+1}, \ldots, a_{t+H-1}\right).
\end{equation}
The sequence of desired configurations forms a reference trajectory for the robot over the planning horizon. We represent the planning policy as $\mathbf{a}_t \sim\pi_\theta(\cdot \mid o_t)$ and use $f_\theta(o_t)$ to denote one $H$-step action sequence sampled from this policy.

\paragraph{Behavior Cloning.}
Let $\mathcal{D}=\{\tau_i\}_{i=1}^{N}$ denote a dataset of $N$ demonstration trajectories, where $\tau_i=\{(o_t,a_t)\}_{t=0}^{T_i-1}$. From each trajectory, we construct valid training pairs consisting of an observation $o_t$ and its corresponding $H$-step future action sequence $\mathbf{a}_t$, and write $(o_t,\mathbf{a}_t)\sim\mathcal{D}$ to denote sampling such a pair from the demonstration dataset. BC trains the policy to reproduce the demonstrated action sequences by minimizing
\begin{equation}
    \mathcal{L}_{\mathrm{BC}}(\theta)
    =
    \mathbb{E}_{(o_t,\mathbf{a}_t)\sim\mathcal{D}}
    \left[
        \ell\left(\pi_\theta(\cdot \mid o_t), \mathbf{a}_t\right)
    \right],
    \label{eq:bc}
\end{equation}
where $\ell$ denotes the training objective associated with the chosen policy model.

\paragraph{Covariate Shift in Closed-Loop Execution.}
Let $d_E(o)$ denote the observation distribution represented by the demonstrations and $d_{\pi_\theta}(o)$ the observation distribution induced by executing the learned policy. BC minimizes its training objective under $d_E$, whereas the policy is evaluated under $d_{\pi_\theta}$ during closed-loop execution. To distinguish the predicted action sequence from the trajectory resulting from its execution, let $\boldsymbol{\tau}_\theta(o)=\Phi\bigl(o,f_\theta(o)\bigr)$, where $\Phi$ denotes the execution mapping from the current observation and predicted action sequence to the resulting robot trajectory. Prediction or execution errors can cause the realized trajectory to deviate from the behaviors represented in the demonstrations, thereby altering subsequent observations. As a result, in general, $d_{\pi_\theta}\neq d_E$. As execution progresses, these deviations can accumulate and drive the robot toward observations that are poorly covered by the demonstrations, where the learned policy may produce increasingly unreliable predictions \citep{ross2011reduction}. This mismatch between the training and execution observation distributions is commonly referred to as \emph{covariate shift}. 

\begin{figure}[tb!]
\begin{center}
\includegraphics[width=0.9\linewidth]{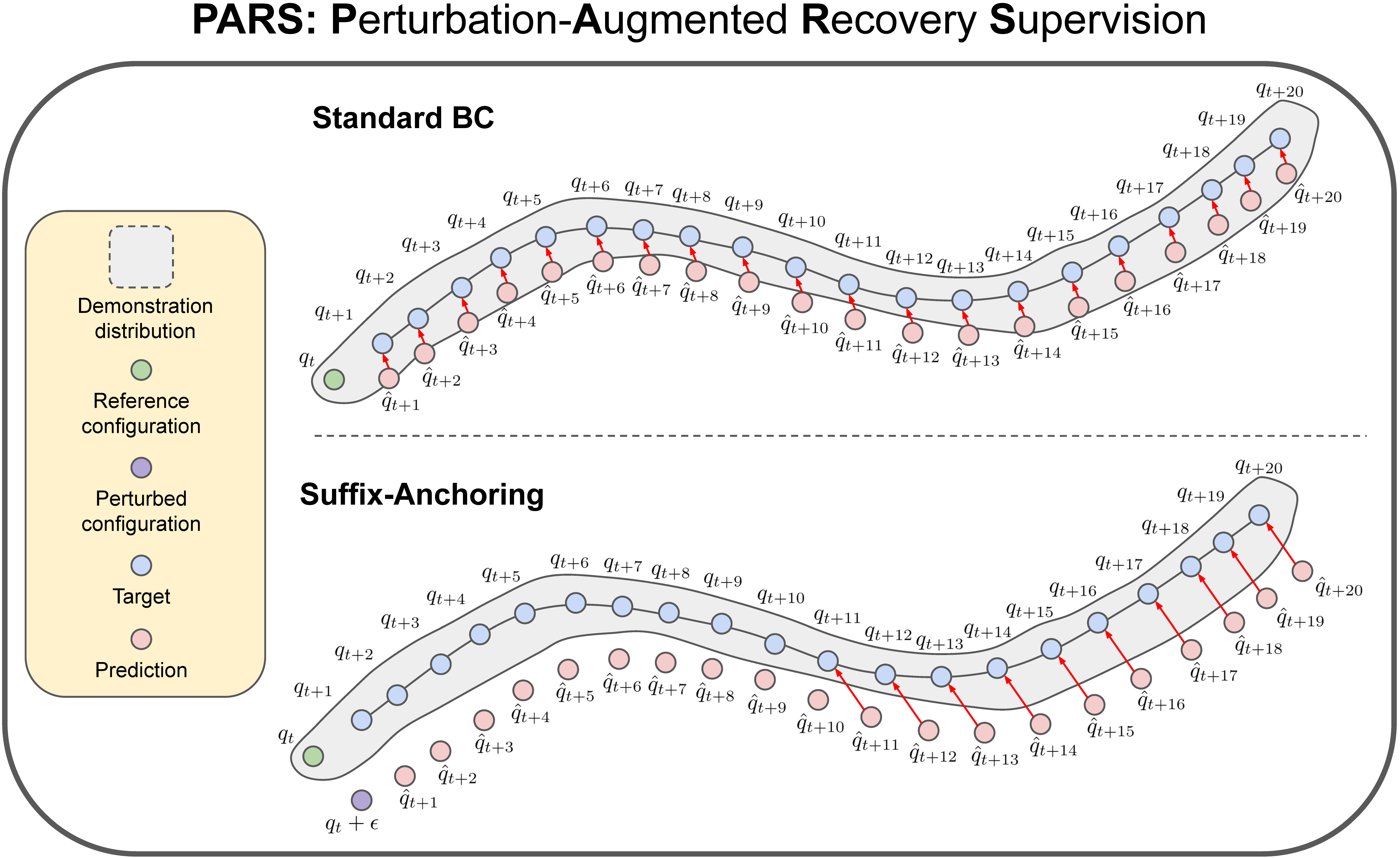}
\end{center}
\caption{Standard BC supervises the full predicted trajectory at an unperturbed observation (top). For a perturbed proprioceptive observation, PARS anchors only the predicted suffix to the corresponding demonstrated suffix (bottom). The prefix receives no direct supervision from the perturbed sample, leaving room for corrective motion.}
\label{fig:PARS}
\end{figure}

\section{Method}
\subsection{Perturbation-Augmented Recovery Supervision}
\label{sec:pars}
We propose Perturbation-Augmented Recovery Supervision (PARS), a simple supervision scheme for improving recovery from covariate shift in action-sequence planning. Under covariate shift, where the robot's proprioceptive state deviates from those represented in the demonstrations, the policy may no longer reliably reproduce the demonstrated motion. The key intuition behind PARS is to encourage the planner to generate a trajectory that approaches the demonstrated trajectory within the planning horizon, bringing the robot toward states where it can more reliably reproduce demonstrated behavior. To induce such recovery behavior, we perturb the proprioceptive observation and apply direct supervision only to the suffix of the predicted action sequence, encouraging it to match the corresponding suffix of the original demonstration, as illustrated in~\cref{fig:PARS}. Formally, we define
\begin{equation}
\begin{aligned}
    \epsilon
    &\sim
    \mathcal{U}\left(-\delta_e, \delta_e\right), \\
    \tilde{o}_t
    &=
    \left(
        o_t^{\mathrm{prop}} + \epsilon,
        o_t^{\mathrm{env}}
    \right), \\
    \left[\mathbf{x}\right]_{H_r:H}
    &=
    \left(
        x_{H_r},
        \ldots,
        x_{H-1}
    \right),
\end{aligned}
\label{eq:pars_def}
\end{equation}
where $\mathcal{U}(-\delta_e,\delta_e)$ denotes a uniform distribution over the task-dependent perturbation range $[-\delta_e,\delta_e]$, and $H_r$ denotes the number of initial planning steps left free from direct augmented supervision. The perturbation $\epsilon$ has dimension $d_e=\dim(o_t^{\mathrm{prop}})$ and is applied to both the robot configuration and gripper state. We write $[\,\cdot\,]_{H_r:H}$ for the suffix of any length-$H$ sequence, so that $[\mathbf{a}_t]_{H_r:H}$ is the demonstrated suffix and $[f_\theta(\tilde{o}_t)]_{H_r:H}$ the corresponding predicted one.

Based on this formulation, for policies that directly predict an action sequence, we augment the standard BC objective with a suffix-anchoring loss:
\begin{equation}
\mathcal{L}_{\mathrm{PARS}}(\theta)
=
\mathbb{E}_{(o_t,\mathbf{a}_t)\sim\mathcal{D}}
\Big[\underbrace{
    \ell\left(
        f_\theta(o_t),
        \mathbf{a}_t
    \right)
    }_{\mathcal{L}_{\mathrm{BC}}}+
    \lambda_{\mathrm{aug}}
    \underbrace{
    \ell\left(
        \left[f_\theta(\tilde{o}_t)\right]_{H_r:H},
        [\mathbf{a}_t]_{H_r:H}
    \right)
    }_{\mathcal{L}_{\mathrm{aug}}}
\Big],
\label{eq:pars}
\end{equation}
Here $\lambda_{\mathrm{aug}}$ represents the contribution of the augmented suffix-anchoring loss. In our implementation, augmented supervision is applied with a task-specific perturbation probability $p$; its resulting weight in the training objective is detailed in Appendix~\ref{app:augmentation-settings}. We first analyze the squared-error case. For policy classes whose training objectives are not defined directly on predicted action sequences, such as diffusion- and flow-based policies, we instantiate the same suffix-supervision principle in their respective training objectives, as described in Section~\ref{sec:masking-applicability}.

\subsection{Theoretical Justification}
\label{sec:theory}
\paragraph{The suffix-anchoring loss acts as a selective invariance regularizer.}
Following the slicing notation defined above, $\left[f_\theta(o)\right]_{H_r:H}$ denotes the suffix of the predicted action sequence. For a training sample $(o_t,\mathbf{a}_t)$, we consider the augmented loss in squared-error form for the following analysis:
\begin{equation}
    \mathcal{L}_{\mathrm{aug}}(\epsilon)
    =
    \frac{1}{2}
    \left\|
        \left[f_\theta(\tilde{o}_t)\right]_{H_r:H}
        -
        \left[\mathbf{a}_t\right]_{H_r:H}
    \right\|^2,
\end{equation}
where $\tilde{o}_t$ is the perturbed observation defined in \cref{eq:pars_def}. This objective compares only the suffix of the action sequence predicted from the perturbed observation with the corresponding suffix of the demonstrated action sequence.

For small perturbations and a small nominal suffix-prediction residual, the expansion in Appendix~\ref{app:jacobian} approximates the expected augmented loss as
\begin{equation}
\mathbb{E}_{\epsilon}
\left[
    \mathcal{L}_{\mathrm{aug}}(\epsilon)
\right]
\approx
\underbrace{
\frac{1}{2}
\left\|
    \left[f_\theta(o_t)\right]_{H_r:H}
    -
    \left[\mathbf{a}_t\right]_{H_r:H}
\right\|^2
}_{\text{unperturbed suffix loss}=\mathcal{L}_{\mathrm{aug}}(0)}
+
\underbrace{
\frac{1}{2}
\operatorname{tr}
\left(
    J_{\mathrm{suf}}
    \Sigma_\epsilon
    J_{\mathrm{suf}}^\top
\right)
}_{\text{suffix sensitivity regularizer}},
\label{eq:theory-jac}
\end{equation}
where
\begin{equation}
    J_{\mathrm{suf}}
    =
    \left.
    \frac{
        \partial [f_\theta(o)]_{H_r:H}
    }{
        \partial o^{\mathrm{prop}}
    }
    \right|_{o=o_t},
\end{equation}
and $\Sigma_\epsilon=\operatorname{Cov}(\epsilon)$. The second term in \cref{eq:theory-jac} penalizes the sensitivity of the predicted suffix to proprioceptive perturbations, while no such direct penalty is imposed on the prefix. Thus, the augmented suffix-anchoring loss acts as a selective invariance regularizer over the later portion of the predicted action sequence. The detailed derivation is provided in Appendix~\ref{app:jacobian}.

\paragraph{Local feasibility and consistency of the free prefix.}
The augmented loss does not directly supervise the prefix predicted from a perturbed observation. We therefore ask whether this prefix stays close to the configuration trajectories in the demonstrations and whether its joint-space speed and acceleration remain within the limits used to generate those demonstrations. We derive sufficient conditions for both properties.

Let $\mathbf{q}^{\mathrm{pre}}_\theta(o)$ denote the desired-configuration prefix generated by the policy, and let $\mathcal{S}^{\mathrm{pre}}_{\mathrm{data}}$ denote the set of configuration prefixes represented in the demonstrations. We define the trajectory-level distance between two prefixes as $d_{\mathrm{traj}}(\mathbf q,\mathbf q')=\max_{0\le k<H_r}\|q_k-q'_k\|_2$ and use $\operatorname{dist}(\cdot,\mathcal{S}^{\mathrm{pre}}_{\mathrm{data}})$ for the distance induced by this metric to the demonstration-prefix set. We assume that the nominal prefix lies within distance $\eta$ of this set and that the predicted prefix is locally Lipschitz with respect to the proprioceptive perturbation:
\begin{equation}
\operatorname{dist}
\left(
    \mathbf{q}^{\mathrm{pre}}_\theta(o_t),
    \mathcal{S}^{\mathrm{pre}}_{\mathrm{data}}
\right)
\le \eta,
\qquad
d_{\mathrm{traj}}
\left(
    \mathbf{q}^{\mathrm{pre}}_\theta(\tilde{o}_t),
    \mathbf{q}^{\mathrm{pre}}_\theta(o_t)
\right)
\le
L_q\|\epsilon\|_2 .
\label{eq:prefix_assumptions}
\end{equation}
Since $\|\epsilon\|_2\le\sqrt{d_e}\,\delta_e$, these assumptions imply
\begin{equation}
\operatorname{dist}
\left(
    \mathbf{q}^{\mathrm{pre}}_\theta(\tilde{o}_t),
    \mathcal{S}^{\mathrm{pre}}_{\mathrm{data}}
\right)
\le
\eta + L_q\sqrt{d_e}\,\delta_e .
\label{eq:consistency_bound}
\end{equation}
Thus, a locally perturbed prefix may deviate from the nominal trajectory while remaining close to demonstrated motion. We additionally assume that the nominal prefix satisfies the joint-space speed and acceleration limits used for trajectory planning with positive margins $m_v$ and $m_a$. Under the same local-sensitivity assumption, the perturbed prefix remains within these limits whenever
\begin{equation}
\frac{2L_q\sqrt{d_e}\,\delta_e}{\Delta t}
\le m_v,
\qquad
\frac{4L_q\sqrt{d_e}\,\delta_e}{\Delta t^2}
\le m_a .
\label{eq:feasibility_condition}
\end{equation}
Together, \cref{eq:consistency_bound,eq:feasibility_condition} show that, under sufficiently small local perturbations, the free prefix can retain flexibility for corrective motion while remaining locally consistent with demonstrated motion and within the joint-space speed and acceleration limits used for trajectory planning. Detailed derivations are provided in Appendix~\ref{app:prefix_analysis}.

\subsection{Instantiating PARS across policy classes}
\label{sec:masking-applicability}
PARS requires the augmented objective to supervise only the suffix of the predicted action sequence. For policies that directly apply a loss to individual steps of the predicted action sequence, this can be implemented by simply restricting the augmented loss to the suffix. However, other policy classes, such as flow-matching and diffusion policies, supervise intermediate generative quantities rather than the final predicted action sequence, requiring different implementations of the same suffix-only supervision. We describe below how PARS is instantiated across the policy classes used in our experiments. The corresponding sensitivity analyses are provided in
Appendix~\ref{app:policy-sensitivity}.

\paragraph{Case 1: direct chunk-prediction policies.}
For policies that directly predict an entire action sequence and compute the loss separately at each step of the sequence, PARS can be directly applied by computing the augmented loss only over the trajectory suffix. The per-step loss is computed on accumulated joint positions, preserving the corrective displacement in the supervised suffix. The transformer policy follows this setting and therefore requires no further modification beyond restricting the augmented loss to the suffix. ACT \citep{zhao2023learning} likewise predicts an entire action chunk, but is trained as a conditional VAE with an $L_1$ reconstruction term and a KL regularization term. For ACT, we apply the suffix restriction to the reconstruction term in the same accumulated trajectory space while leaving the KL term unchanged.

\paragraph{Case 2: streaming flow policies.}
Streaming Flow Policy (SFP) \citep{pmlr-v305-jiang25a} treats the action trajectory itself as a flow trajectory: integrating its learned velocity field advances the predicted action along sequence time $\tau\in[0,1]$. For PARS, we apply the augmented velocity-regression loss only after $\tau_r:=H_r/H$. Supervising velocities over this suffix does not fix the action reached at $\tau_r$, which depends on the freely generated prefix. We therefore integrate the policy's velocity field up to $\tau_r$ for each perturbed observation and use the resulting action as the starting point for suffix supervision. The remaining sequence is then trained toward the demonstrated continuation, while the prefix receives no direct augmented supervision.

\paragraph{Case 3: joint generative policies.}
Conventional flow-matching and diffusion policies do not directly regress the target action sequence. Instead, they learn a velocity or noise field from complete target sequences, making suffix-only supervision less straightforward. Under a perturbed observation $\tilde{o}_t$, the corresponding recovery trajectory is unavailable in the demonstration dataset: the desired recovery prefix is unknown, while the available supervision specifies only that the later portion of the trajectory should remain consistent with the demonstrated suffix. For a perturbed observation, we generate an action sequence with the current FP or DP policy and refine it by reducing the suffix-anchoring loss on accumulated joint configurations. The refined sequence then serves as a surrogate target for the policy's original flow-matching or diffusion training objective. This adapts the sample-refinement principle of \citet{park2026refinement} to suffix supervision; the refinement settings are reported in Appendix~\ref{app:augmentation-settings}.

\section{Experiments}
\subsection{Experimental Setup}
\label{sec:experimental-setup}
We evaluate PARS on 51 RLBench manipulation tasks \citep{james2020rlbench} using a Franka Panda 7-DoF robotic arm. We use low-dimensional state observations rather than images. Each observation contains the robot's joint positions, joint velocities, and gripper state, together with task-specific low-dimensional information. Given the current observation, each policy predicts a 20-step action sequence. Each action is represented as $[\Delta q,g]$, where $\Delta q\in\mathbb{R}^{7}$ is a joint-space displacement and $g$ is a gripper command. We use RLBench's \texttt{JointPosition} action mode with absolute joint-position targets. The predicted displacements are accumulated from the robot configuration at the beginning of the sequence to obtain these targets, while $g$ is thresholded at $0.5$ to issue a binary open/close command. For each task, we collect 100 expert demonstrations of a single task variation using RLBench's built-in demonstration generator and construct training pairs from each observation and its corresponding 20-step future action sequence. During evaluation, all 20 predicted actions are executed sequentially without re-observation or replanning. The policy then receives a new observation and predicts the next sequence. We report task success rate (\%) over 50 evaluation episodes per task, each limited to 300 environment steps.

To evaluate PARS across action-sequence policy formulations, we consider three representative classes: flow-based, transformer-based, and diffusion-based policies. Their standard baselines are a flow-matching policy (FP) \citep{lipman2023flow}, a transformer policy (TP) \citep{vaswani2017attention}, and diffusion policy (DP) \citep{chi2025diffusion}, respectively. We also include Action Chunking with Transformers (ACT) \citep{zhao2023learning}, which models action sequences using a conditional VAE. To examine whether PARS complements existing robustness mechanisms, we evaluate Streaming Flow Policy (SFP) \citep{pmlr-v305-jiang25a}, which learns a stabilizing flow field around demonstrated trajectories, and KDPE \citep{rosasco2025kdpe}, which uses kernel density estimation to select a high-density action sequence from multiple DP candidates. Each trainable policy uses its respective BC objective, with the PARS instantiation described in Section~\ref{sec:masking-applicability}. KDPE is applied only at inference time to the corresponding DP checkpoint.

\begin{table}[t]
\centering
\caption{Performance of PARS across different action-sequence methods on 51 RLBench tasks. Mean success rate (\%) over 50 evaluation episodes; for PARS, we report the best per-task augmentation setting. The last column counts the number of tasks on which PARS matches or exceeds the corresponding baseline. Full results for all 51 tasks are provided in Table~\ref{tab:pertask}.}
\label{tab:main}

\begin{tabular}{llcccc}
\hline
&
\multirow{2}{*}{Method}
& \multicolumn{2}{c}{PARS Comparison}
& \multirow{2}{*}{\shortstack{Noise\\Augmentation}}
& \multirow{2}{*}{\shortstack{Tasks w/ PARS\\$\geq$ w/o PARS}} \\
\cline{3-4}
&
& w/o
& w/
&
& \\
\hline

\multirow{2}{*}{\textbf{Flow-based}}
& FP   & 58.0 & \textbf{66.9} & 60.0 & 48/51 \\
& SFP  & 49.8 & \textbf{61.6} & 46.7 & 44/51 \\
\hline

\multirow{2}{*}{\textbf{Transformer-based}}
& TP   & 57.4 & \textbf{67.7} & 59.3 & 46/51 \\
& ACT  & 58.8 & \textbf{68.2} & 60.0 & 48/51 \\
\hline

\multirow{2}{*}{\textbf{Diffusion-based}}
& DP   & 47.6 & \textbf{56.0} & 46.2 & 46/51 \\
& KDPE & 53.5 & \textbf{60.2} & 56.7 & 42/51 \\
\hline

\end{tabular}
\end{table}

% The experiments are designed to answer the following questions.
% \begin{itemize}
%     \item Does PARS improve performance during closed-loop execution across different action-sequence methods?
    
%     \item How does PARS differ from conventional input-noise augmentation, and how does this difference affect recovery during closed-loop execution?
    
%     \item Does PARS promote recovery toward demonstrated behavior over the prediction horizon?

%     \item Does the free recovery prefix remain locally consistent with demonstrated motion while satisfying the trajectory-planning limits?
% \end{itemize}

\subsection{Experimental Results}
\paragraph{Improvement of closed-loop performance across different action-sequence methods.}
Table~\ref{tab:main} compares task success rates with and without PARS across flow-based, transformer-based, and diffusion-based action-sequence policies. PARS improves the mean success rate for all evaluated methods and matches or exceeds the corresponding baseline on most tasks. These improvements span different policy formulations, suggesting that the benefits of PARS are not limited to a particular architecture or training objective. PARS also improves performance when combined with SFP and KDPE, suggesting that it can complement existing methods for improving robustness during execution. Overall, these results support the applicability of PARS across diverse action-sequence methods.

\paragraph{Benefit of Suffix Anchoring over Full-Target Supervision.}
We compare PARS with conventional noise augmentation to examine how supervision of perturbed observations affects closed-loop success. Both perturb proprioceptive observations during training, but conventional noise augmentation retains the original demonstrated action sequence as the target over the full prediction horizon, encouraging consistent predictions under perturbations. PARS instead anchors only the suffix to the demonstrated continuation, leaving the prefix free to produce corrective motion. As shown in Table~\ref{tab:main}, conventional noise augmentation yields modest gains or degrades performance, whereas PARS improves mean success rates for all six policy methods. These results suggest that suffix anchoring yields better closed-loop performance than full-target supervision under input perturbations.

\begin{figure}[tb!]
\begin{center}
\includegraphics[width=\linewidth]{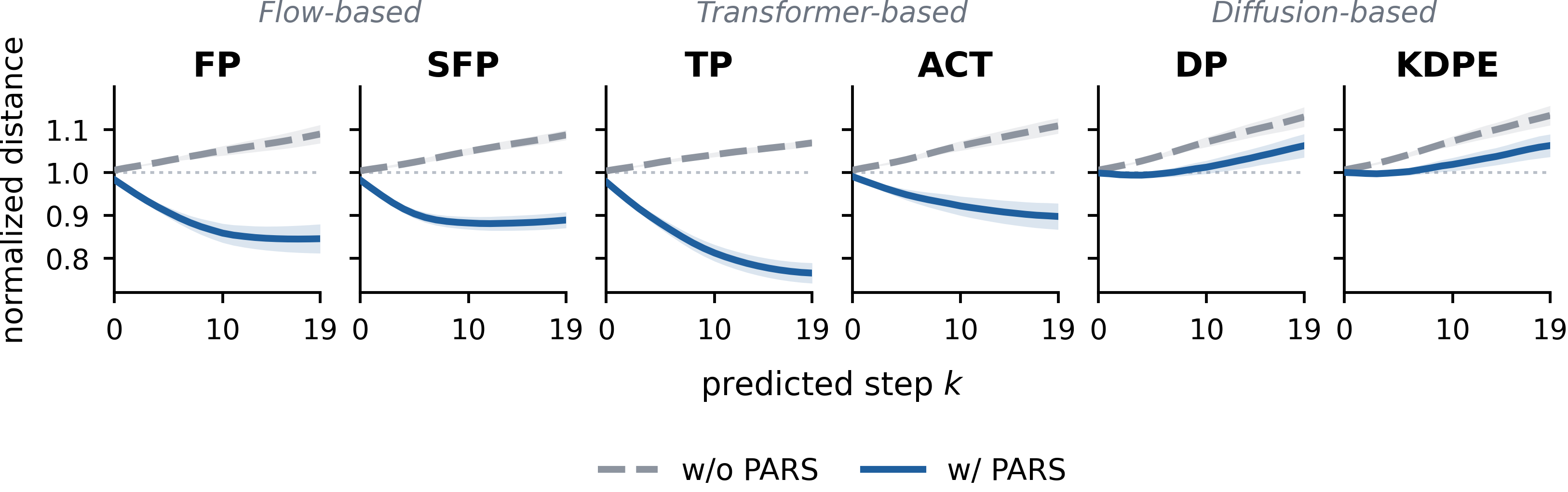}
\end{center}
\caption{Normalized distance to the demonstrated configurations over the predicted action sequence after deviation. Values below $1.0$ indicate recovery toward the demonstrations, whereas values above $1.0$ indicate further deviation. Curves show the mean over 51 RLBench tasks with $\pm1$ standard error. PARS promotes recovery or reduces divergence across all evaluated policies.}
\label{fig:recovery}
\end{figure}
\paragraph{Recovery toward the Demonstrated Trajectory.}
To examine whether PARS's performance gains are associated with recovery, we evaluate predicted sequences at off-nominal observations reached by the unaugmented policies. Figure~\ref{fig:recovery} shows, over predicted step $k$, the nearest-neighbor distance from each predicted joint configuration to the joint configurations in that task's demonstrations, normalized by its value at $k=0$. The first $H_r=10$ steps form the free recovery prefix and receive no direct augmented supervision. Without PARS, distances generally increase over the horizon. With PARS, the distances end below their initial levels for FP, SFP, TP, and ACT, although the FP and SFP curves flatten or rise slightly near the end. For DP and KDPE, the distances end above their initial levels, but PARS substantially reduces the final divergence relative to the unaugmented policies.

\begin{figure}[t]
\begin{center}
\includegraphics[width=\linewidth]{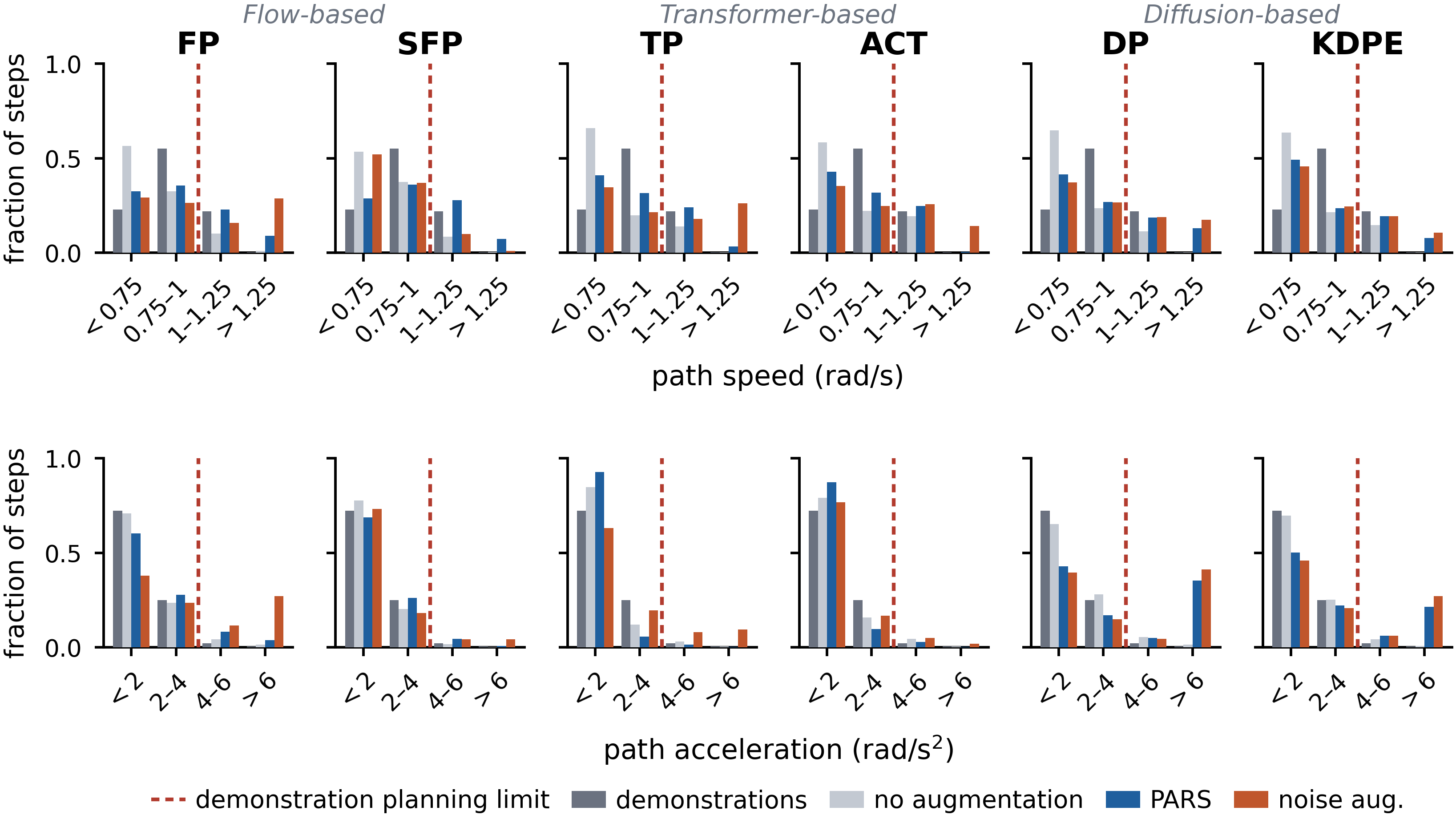}
\end{center}
\caption{\textbf{Kinematics of the free recovery prefix.}
Distributions of path speed ($\lVert\Delta q\rVert_2/\Delta t$) and path acceleration ($\lVert\Delta^2 q\rVert_2/\Delta t^2$) over the first $H_r=10$ predicted steps. Bars show the fraction of steps in each range, averaged over 51 RLBench tasks. Dashed lines indicate the RLBench planning limits.}
\label{fig:prefix_kinematics}
\end{figure}

\paragraph{Kinematic behavior of the free recovery prefix.}
We examine whether leaving the recovery prefix without direct augmented supervision leads to excessively aggressive motion. Figure~\ref{fig:prefix_kinematics} compares the path-speed and path-acceleration distributions of the free prefix for the demonstrations, unaugmented policies, PARS, and conventional noise augmentation. Dashed lines indicate the joint-space speed and acceleration limits used to plan the demonstrations. Across most policy classes, PARS places less probability mass at high speeds and accelerations than noise augmentation, with particularly clear differences for FP, TP, and ACT. SFP also remains largely within the demonstrated motion range. For DP and KDPE, PARS reduces the high-acceleration tail relative to noise augmentation, although a noticeable fraction of predicted steps still exceeds the demonstration planning limit for acceleration. Thus, leaving the prefix free from direct augmented supervision does not necessarily produce aggressive motion and, in most cases, yields kinematics closer to those of the demonstrations than noise augmentation. Our feasibility analysis is conditional on the stated local-sensitivity and margin assumptions and therefore does not guarantee that every predicted prefix satisfies the planning limits.

\section{Conclusion}
We presented Perturbation-Augmented Recovery Supervision (PARS), a supervision scheme for improving recovery from covariate shift in action-sequence BC using existing demonstrations. PARS perturbs the proprioceptive observation and applies direct augmented supervision only to the suffix of the predicted action sequence, leaving the preceding prefix room for corrective motion. Our analysis characterizes the local sensitivity induced by suffix anchoring and derives conditions under which the free prefix remains close to demonstrated motion and satisfies the joint-space speed and acceleration limits used to plan the demonstrations. Across 51 RLBench tasks, PARS increases mean closed-loop success for all six evaluated policy methods and outperforms conventional full-target noise augmentation in mean success. Predicted trajectories approach demonstrated configurations for FP, SFP, TP, and ACT, while PARS reduces further divergence for DP and KDPE. These results suggest that recovery behavior can be encouraged using existing demonstrations without collecting explicit corrective trajectories or querying an expert during training.

\paragraph{Limitations.}
PARS uses task-specific perturbation settings that do not adapt to the stage of an episode or to individual proprioceptive dimensions. Our goal in this work is to demonstrate that perturbing proprioceptive observations in existing demonstrations and anchoring the predicted suffix can improve recovery, rather than to identify an optimal perturbation scheme for each task. Adaptive perturbations may better reflect how recovery demands change with context. We also use a fixed free recovery prefix of $H_r=10$ steps within a 20-step prediction horizon. Its appropriate length may depend on the task and perturbation magnitude; larger deviations may require more steps before the predicted trajectory can approach demonstrated motion. Finally, our evaluation uses simulated RLBench tasks with low-dimensional observations and one variation per task. Extending the evaluation to visual observations and physical robots remains future work.

\section*{AI use statement}
In this work, we used generative AI tools to refine conceptual framing, provide feedback on research methodology and experimental design, and assist with translation of technical descriptions. We have not used generative AI tools to generate synthetic datasets, formulate mathematical claims, propose or refine hypotheses, implement methods, clean or reformat datasets, or interpret experimental results, and proof-related tasks and qualitative or thematic data analysis are not applicable to this work. Additionally, we used generative AI tools for brainstorming, searching for and identifying relevant literature, summarizing existing literature, suggesting the structure of the manuscript, drafting parts of the manuscript, and editing the manuscript to improve clarity and readability. We have reviewed all AI-assisted work. All AI-assisted methodological suggestions and manuscript text were critically reviewed and revised by the authors, and cited literature and technical claims were independently checked against the original sources. We take responsibility for the final content of this work, including text, claims or artifacts produced with the aid of generative AI.

\subsection*{Reproducibility statement}
Sections~\ref{sec:pars} and~\ref{sec:masking-applicability} define PARS and describe its instantiation across policy classes. Section~\ref{sec:theory} and Appendices~\ref{app:policy-sensitivity} and~\ref{app:prefix_analysis} state the assumptions and provide the theoretical derivations. Section~\ref{sec:experimental-setup} describes the RLBench tasks, demonstration collection, action representation, and evaluation protocol. Appendices~\ref{app:training-details}--\ref{app:best-settings} report the training settings, selected task-specific PARS settings, and protocols for the recovery and kinematic analyses. Table~\ref{tab:pertask} reports success rates for all 51 tasks.

% Use unnumbered third level headings for the acknowledgments. All acknowledgments, including those to funding agencies, go at the end of the paper.

\bibliography{iclr2027_conference}
\bibliographystyle{iclr2027_conference}

\clearpage
\appendix

\section{Why We Focus on Action-Sequence Policies}
\label{app:single-step}
Single-step behavior cloning is known to be particularly susceptible to compounding errors in closed-loop execution, where small prediction errors can shift the policy away from the state distribution covered by expert demonstrations \citep{ross2011reduction}. Recent studies further show that action-sequence prediction can substantially improve behavior-cloning performance by reducing compounding errors and exploiting temporal information across multiple actions \citep{lazzati2026does, zhang2026action}.

Consistent with these observations, the single-step BC-based baselines showed limited closed-loop performance on our manipulation tasks, as summarized in Table~\ref{tab:single}. Because these methods exhibit low closed-loop performance in our setting, we focus the main evaluation on action-sequence policies, which provide stronger baselines for evaluating the recovery and robustness improvements introduced by PARS.
\begin{table}[h]
\centering
\caption{Closed-loop success rates (\%) of single-step BC-based baselines, averaged over 51 tasks.}
\label{tab:single}
\begin{tabular}{cccccc}
\hline
Vanilla BC & CCIL & DrilDICE & MBIL & KATS & Stable-BC \\
\hline
0.5 & 1.2 & 1.6 & 0.8 & 1.7 & 9.8 \\
\hline
\end{tabular}
\end{table}

\section{Sensitivity Analyses across Policy Classes}
\label{app:policy-sensitivity}
Section~\ref{sec:masking-applicability} describes how the suffix-only supervision of PARS is instantiated across the policy classes considered in our experiments. We provide here the corresponding local sensitivity analyses. The exact form of the induced regularization depends on the quantity supervised by each training objective. Direct trajectory regression admits the Jacobian regularization derived in Section~\ref{sec:theory}, whereas objectives based on $L_1$ reconstruction or generative-field regression require separate extensions.

\subsection{Direct Chunk Prediction with a Squared Loss}
\label{app:jacobian}
We first derive the approximation in \cref{eq:theory-jac}, which applies directly to chunk-prediction policies trained with a squared trajectory loss. For a demonstration sample $(o_t,\mathbf{a}_t)$, PARS perturbs only the proprioceptive component of the observation:
\begin{equation}
    \tilde{o}_t
    =
    \left(
        o_t^{\mathrm{prop}}+\epsilon,
        o_t^{\mathrm{env}}
    \right),
\end{equation}
where
\begin{equation}
    \mathbb{E}[\epsilon]=0,
    \qquad
    \Sigma_\epsilon
    =
    \operatorname{Cov}(\epsilon).
\end{equation}

The augmented loss considered in the analysis is
\begin{equation}
    \mathcal{L}_{\mathrm{aug}}(\epsilon)
    =
    \frac{1}{2}
    \left\|
        [f_\theta(\tilde{o}_t)]_{H_r:H}
        -
        [\mathbf{a}_t]_{H_r:H}
    \right\|^2.
\label{eq:app-direct-loss}
\end{equation}

Assuming that $f_\theta$ is sufficiently smooth around $o_t$, a second-order Taylor expansion with respect to the proprioceptive input gives
\begin{equation}
\begin{aligned}
    [f_\theta(\tilde{o}_t)]_{H_r:H}
    =
    &[f_\theta(o_t)]_{H_r:H}
    +
    J_{\mathrm{suf}}\epsilon
    \\
    &+
    \frac{1}{2}
    Q_{\mathrm{suf}}(\epsilon,\epsilon)
    +
    O(\|\epsilon\|^3),
\end{aligned}
\label{eq:app-direct-taylor}
\end{equation}
where
\begin{equation}
    J_{\mathrm{suf}}
    =
    \left.
    \frac{
        \partial [f_\theta(o)]_{H_r:H}
    }{
        \partial o^{\mathrm{prop}}
    }
    \right|_{o=o_t},
\end{equation}
and $Q_{\mathrm{suf}}(\epsilon,\epsilon)$ collects the second-order curvature terms of the predicted suffix.

Let
\begin{equation}
    r_t
    =
    [f_\theta(o_t)]_{H_r:H}
    -
    [\mathbf{a}_t]_{H_r:H}
\end{equation}
denote the suffix prediction residual at the original observation. Substituting \cref{eq:app-direct-taylor} into \cref{eq:app-direct-loss} and retaining terms up to second order yields
\begin{equation}
\begin{aligned}
\mathcal{L}_{\mathrm{aug}}(\epsilon)
=
&\frac{1}{2}\|r_t\|^2
+
r_t^\top J_{\mathrm{suf}}\epsilon
\\
&+
\frac{1}{2}
\epsilon^\top
J_{\mathrm{suf}}^\top
J_{\mathrm{suf}}
\epsilon
\\
&+
\frac{1}{2}
r_t^\top
Q_{\mathrm{suf}}(\epsilon,\epsilon)
+
O(\|\epsilon\|^3).
\end{aligned}
\label{eq:app-direct-expand}
\end{equation}

Taking expectation over the perturbation eliminates the first-order term because $\mathbb{E}[\epsilon]=0$. Moreover,
\begin{equation}
    \mathbb{E}_{\epsilon}
    \left[
        \epsilon^\top
        J_{\mathrm{suf}}^\top
        J_{\mathrm{suf}}
        \epsilon
    \right]
    =
    \operatorname{tr}
    \left(
        J_{\mathrm{suf}}
        \Sigma_\epsilon
        J_{\mathrm{suf}}^\top
    \right).
\end{equation}
Hence,
\begin{equation}
\begin{aligned}
\mathbb{E}_{\epsilon}
\left[
    \mathcal{L}_{\mathrm{aug}}(\epsilon)
\right]
=
&\frac{1}{2}\|r_t\|^2
+
\frac{1}{2}
\operatorname{tr}
\left(
    J_{\mathrm{suf}}
    \Sigma_\epsilon
    J_{\mathrm{suf}}^\top
\right)
\\
&+
\frac{1}{2}
r_t^\top
\mathbb{E}_{\epsilon}
\left[
    Q_{\mathrm{suf}}(\epsilon,\epsilon)
\right]
+
O\!\left(
    \mathbb{E}_{\epsilon}\|\epsilon\|^3
\right).
\end{aligned}
\label{eq:app-direct-full}
\end{equation}

When the trained policy closely matches the demonstrated suffix, $r_t$ is small. For sufficiently small perturbations, the residual-dependent curvature term and higher-order terms can therefore be neglected, giving
\begin{equation}
\mathbb{E}_{\epsilon}
\left[
    \mathcal{L}_{\mathrm{aug}}(\epsilon)
\right]
\approx
\frac{1}{2}\|r_t\|^2
+
\frac{1}{2}
\operatorname{tr}
\left(
    J_{\mathrm{suf}}
    \Sigma_\epsilon
    J_{\mathrm{suf}}^\top
\right),
\end{equation}
which gives \cref{eq:theory-jac}.

For independent uniform perturbations $\epsilon_i\sim\mathcal{U}(-\delta_e,\delta_e)$,
\begin{equation}
    \Sigma_\epsilon
    =
    \frac{\delta_e^2}{3}I,
\end{equation}
and therefore
\begin{equation}
    \frac{1}{2}
    \operatorname{tr}
    \left(
        J_{\mathrm{suf}}
        \Sigma_\epsilon
        J_{\mathrm{suf}}^\top
    \right)
    =
    \frac{\delta_e^2}{6}
    \|J_{\mathrm{suf}}\|_F^2.
\end{equation}

Thus, the augmented squared loss locally penalizes the sensitivity of the predicted suffix to proprioceptive perturbations. Because the augmented loss contains only the trajectory suffix, no corresponding Jacobian penalty is directly introduced for the trajectory prefix. This analysis directly covers the squared-loss chunk-prediction policy considered in our experiments.

\subsection{Direct Chunk Prediction with an $L_1$ Loss}
\label{app:l1}
We next consider a direct chunk-prediction policy trained using an $L_1$ reconstruction loss, as in ACT. Only the reconstruction term is affected by the PARS suffix restriction; the KL regularization term is left unchanged.

Using the same suffix residual $r_t$ and Jacobian $J_{\mathrm{suf}}$ defined above, a first-order Taylor expansion gives
\begin{equation}
    [f_\theta(\tilde{o}_t)]_{H_r:H}
    =
    [f_\theta(o_t)]_{H_r:H}
    +
    J_{\mathrm{suf}}\epsilon
    +
    O(\|\epsilon\|^2).
\end{equation}
The augmented $L_1$ reconstruction loss is therefore
\begin{equation}
    \mathcal{L}_{\mathrm{aug}}^{L_1}(\epsilon)
    =
    \left\|
        r_t
        +
        J_{\mathrm{suf}}\epsilon
        +
        O(\|\epsilon\|^2)
    \right\|_1.
\end{equation}

In the near-zero-residual regime, $r_t\approx 0$. For sufficiently small perturbations, neglecting the residual and higher-order terms gives
\begin{equation}
    \mathcal{L}_{\mathrm{aug}}^{L_1}(\epsilon)
    \approx
    \left\|
        J_{\mathrm{suf}}\epsilon
    \right\|_1.
\end{equation}
Taking expectation over the perturbation yields
\begin{equation}
\mathbb{E}_{\epsilon}
\left[
    \mathcal{L}_{\mathrm{aug}}^{L_1}(\epsilon)
\right]
\approx
\mathbb{E}_{\epsilon}
\left[
    \left\|
        J_{\mathrm{suf}}\epsilon
    \right\|_1
\right].
\label{eq:app-l1-sensitivity}
\end{equation}

Thus, although the exact form of the induced regularizer differs from the quadratic Jacobian penalty obtained under a squared loss, the augmented $L_1$ reconstruction likewise penalizes the local sensitivity of the supervised suffix to proprioceptive perturbations. The trajectory prefix does not receive an analogous invariance penalty from the augmented reconstruction loss.

\subsection{Streaming Flow Policies}
\label{app:sfp-sensitivity}
We next analyze SFP, for which the augmented objective supervises a velocity field rather than directly supervising the predicted trajectory. Let $v_\theta(a,\tau\mid o)$ denote the observation-conditioned velocity field, where $\tau\in[0,1]$ denotes normalized trajectory time. As described in Section~\ref{sec:masking-applicability}, for perturbed samples the learned flow is first evolved up to $\tau_r$ using the same stepwise integration as at inference, and suffix supervision is then applied from the state reached at $\tau_r$. We analyze below the resulting velocity-regression objective over the supervised suffix.

For clarity, we consider a fixed demonstration trajectory $\xi$ and omit the outer expectation over the demonstration dataset. Let $v_\xi(a,\tau)$ denote the target conditional velocity field associated with $\xi$. The suffix velocity-regression loss can be written as
\begin{equation}
\mathcal{L}_{\mathrm{aug}}^{\mathrm{SFP}}(\epsilon)
=
\mathbb{E}_{\tau,a}
\left[
    \frac{1}{2}
    \left\|
        v_\theta(a,\tau\mid\tilde{o}_t)
        -
        v_\xi(a,\tau)
    \right\|^2
\right],
\qquad
\tau\ge\tau_r,
\label{eq:app-sfp-loss}
\end{equation}
where the expectation is taken over the sampling distribution used by the SFP training objective restricted to the supervised suffix.

Define
\begin{equation}
    r_v(a,\tau)
    =
    v_\theta(a,\tau\mid o_t)
    -
    v_\xi(a,\tau)
\end{equation}
and
\begin{equation}
    J_v(a,\tau)
    =
    \left.
    \frac{
        \partial v_\theta(a,\tau\mid o)
    }{
        \partial o^{\mathrm{prop}}
    }
    \right|_{o=o_t}.
\end{equation}

A second-order Taylor expansion with respect to the proprioceptive perturbation gives
\begin{equation}
\begin{aligned}
v_\theta(a,\tau\mid\tilde{o}_t)
=
&v_\theta(a,\tau\mid o_t)
+
J_v(a,\tau)\epsilon
\\
&+
\frac{1}{2}
Q_v(a,\tau;\epsilon,\epsilon)
+
O(\|\epsilon\|^3),
\end{aligned}
\label{eq:app-sfp-taylor}
\end{equation}
where $Q_v$ collects the second-order curvature terms.

Proceeding as in Appendix~\ref{app:jacobian}, taking expectation over a zero-mean perturbation gives
\begin{equation}
\begin{aligned}
\mathbb{E}_{\epsilon}
\left[
    \mathcal{L}_{\mathrm{aug}}^{\mathrm{SFP}}(\epsilon)
\right]
=
&\mathcal{L}_{\mathrm{aug}}^{\mathrm{SFP}}(0)
\\
&+
\frac{1}{2}
\mathbb{E}_{\tau,a}
\left[
    \operatorname{tr}
    \left(
        J_v
        \Sigma_\epsilon
        J_v^\top
    \right)
\right]
\\
&+
\frac{1}{2}
\mathbb{E}_{\tau,a}
\left[
    r_v^\top
    \mathbb{E}_{\epsilon}
    \left[
        Q_v(\epsilon,\epsilon)
    \right]
\right]
\\
&+
O\!\left(
    \mathbb{E}_{\epsilon}\|\epsilon\|^3
\right).
\end{aligned}
\label{eq:app-sfp-full}
\end{equation}

When the learned velocity field closely matches its target over the supervised suffix, the residual-dependent curvature term becomes small. For sufficiently small perturbations,
\begin{equation}
\begin{aligned}
\mathbb{E}_{\epsilon}
\left[
    \mathcal{L}_{\mathrm{aug}}^{\mathrm{SFP}}(\epsilon)
\right]
\approx\;&
\mathcal{L}_{\mathrm{aug}}^{\mathrm{SFP}}(0)
\\
&+
\frac{1}{2}
\mathbb{E}_{\tau,a}
\left[
    \operatorname{tr}
    \left(
        J_v
        \Sigma_\epsilon
        J_v^\top
    \right)
\right].
\end{aligned}
\label{eq:app-sfp-field-sensitivity}
\end{equation}
Thus, the PARS augmentation locally penalizes the conditioning sensitivity of the velocity field over the supervised suffix.

\paragraph{Relation to trajectory sensitivity.}
Because SFP supervises a velocity field rather than the final trajectory, the field-level regularization above does not directly equal the trajectory Jacobian regularization in \cref{eq:theory-jac}. To characterize their relationship, let $a_o(\tau)$ denote the generated trajectory under observation $o$, satisfying
\begin{equation}
    \frac{d a_o(\tau)}{d\tau}
    =
    v_\theta(a_o(\tau),\tau\mid o).
\label{eq:app-sfp-ode}
\end{equation}
Define its sensitivity to the proprioceptive observation as
\begin{equation}
    S(\tau)
    =
    \frac{
        \partial a_o(\tau)
    }{
        \partial o^{\mathrm{prop}}
    }.
\end{equation}
Differentiating \cref{eq:app-sfp-ode} gives
\begin{equation}
    \frac{dS(\tau)}{d\tau}
    =
    A(\tau)S(\tau)
    +
    B(\tau),
\label{eq:app-sfp-sensitivity-dynamics}
\end{equation}
where
\begin{equation}
    A(\tau)
    =
    \left.
    \frac{
        \partial v_\theta(a,\tau\mid o)
    }{
        \partial a
    }
    \right|_{a=a_o(\tau)}
\end{equation}
and
\begin{equation}
    B(\tau)
    =
    \left.
    \frac{
        \partial v_\theta(a,\tau\mid o)
    }{
        \partial o^{\mathrm{prop}}
    }
    \right|_{a=a_o(\tau)}.
\end{equation}

Let $\Phi(\tau,s)$ denote the state-transition matrix associated with $A(\tau)$. We assume that the learned flow is locally contractive over the relevant suffix region, such that for some $\kappa>0$,
\begin{equation}
    \|\Phi(\tau,s)\|
    \le
    e^{-\kappa(\tau-s)},
    \qquad
    \tau\ge s\ge\tau_r.
\label{eq:app-sfp-contraction}
\end{equation}
This is a local assumption on the learned dynamics and does not imply global contraction of the policy.

The variation-of-constants formula yields
\begin{equation}
    S(\tau)
    =
    \Phi(\tau,\tau_r)S(\tau_r)
    +
    \int_{\tau_r}^{\tau}
    \Phi(\tau,s)B(s)\,ds.
\end{equation}
Taking norms and using \cref{eq:app-sfp-contraction},
\begin{equation}
\begin{aligned}
\|S(\tau)\|
\le\;&
e^{-\kappa(\tau-\tau_r)}
\|S(\tau_r)\|
\\
&+
\int_{\tau_r}^{\tau}
e^{-\kappa(\tau-s)}
\|B(s)\|\,ds.
\end{aligned}
\end{equation}
The integral term satisfies
\begin{equation}
\begin{aligned}
\int_{\tau_r}^{\tau}
e^{-\kappa(\tau-s)}
\|B(s)\|\,ds
\le\;&
\frac{
    1-e^{-\kappa(\tau-\tau_r)}
}{
    \kappa
}
\sup_{s\in[\tau_r,\tau]}
\|B(s)\|.
\end{aligned}
\end{equation}
Since
\begin{equation}
    \frac{
        1-e^{-\kappa(\tau-\tau_r)}
    }{
        \kappa
    }
    \le
    \tau-\tau_r
    \le
    1,
\end{equation}
we obtain
\begin{equation}
\left\|
\frac{\partial a_o(\tau)}{\partial o^{\mathrm{prop}}}
-
\Phi(\tau,\tau_r)
\frac{\partial a_o(\tau_r)}{\partial o^{\mathrm{prop}}}
\right\|
\le
\sup_{s\in[\tau_r,\tau]}
\left\|
\frac{\partial v_\theta(a_o(s),s\mid o)}
{\partial o^{\mathrm{prop}}}
\right\|.
\label{eq:app-sfp-trajectory-sensitivity}
\end{equation}

The second term on the left represents sensitivity inherited from the beginning of the supervised suffix after attenuation by the locally contractive flow. The right-hand side captures the contribution of the observation-conditioned velocity field over the suffix. Therefore, PARS does not directly regularize the complete trajectory-suffix Jacobian in SFP; rather, it regularizes one source of trajectory sensitivity through the velocity field.

This analysis is local. In particular, \cref{eq:app-sfp-field-sensitivity} controls the velocity-field sensitivity under the SFP training distribution, whereas \cref{eq:app-sfp-trajectory-sensitivity} evaluates this sensitivity along a generated trajectory. Their connection therefore assumes that the generated suffix remains in a region where the learned field is locally controlled by the augmented training objective.

\subsection{Joint Generative Policies}
\label{app:joint-generative}
We finally consider joint flow-matching and diffusion policies. Unlike SFP, these models generate the entire action sequence jointly along a generative axis that is distinct from sequence time. For the purpose of analysis, we represent the effect of the refinement-based optimization as inducing a surrogate recovery trajectory $\hat{\mathbf{a}}_t(\epsilon)$ that is consistent with the suffix-level objective.

The following analysis characterizes the corresponding generative-field sensitivity using this surrogate representation. Let $s$ denote the generative time and $\omega$ denote the auxiliary randomness used by the training objective, such as a base sample or sampled noise. Given the surrogate trajectory $\hat{\mathbf{a}}_t(\epsilon)$, let
\begin{equation}
    z_s(\epsilon)
    =
    z_s\left(
        \hat{\mathbf{a}}_t(\epsilon),
        \omega
    \right)
\end{equation}
denote the intermediate noisy or interpolated sample supplied to the model, and let
\begin{equation}
    u_s(\epsilon)
    =
    u_s\left(
        \hat{\mathbf{a}}_t(\epsilon),
        \omega
    \right)
\end{equation}
denote the corresponding velocity, noise, or other regression target.

For compactness, define
\begin{equation}
    \mathcal{F}_\theta(\epsilon)
    =
    F_\theta
    \left(
        z_s(\epsilon),
        s
        \mid
        \tilde{o}_t
    \right)
\end{equation}
and
\begin{equation}
    \mathcal{U}(\epsilon)
    =
    u_s(\epsilon),
\end{equation}
where $F_\theta$ denotes the field predicted by the generative model. The augmented generative loss takes the generic squared form
\begin{equation}
    \mathcal{L}_{\mathrm{aug}}^{\mathrm{gen}}(\epsilon)
    =
    \mathbb{E}_{s,\omega}
    \left[
        \frac{1}{2}
        \left\|
            \mathcal{F}_\theta(\epsilon)
            -
            \mathcal{U}(\epsilon)
        \right\|^2
    \right].
\label{eq:app-gen-loss}
\end{equation}

Define the residual at the unperturbed observation as
\begin{equation}
    r_g
    =
    \mathcal{F}_\theta(0)
    -
    \mathcal{U}(0),
\end{equation}
and define the total perturbation sensitivities
\begin{equation}
    J_F
    =
    \left.
    \frac{
        \partial \mathcal{F}_\theta(\epsilon)
    }{
        \partial \epsilon
    }
    \right|_{\epsilon=0},
    \qquad
    J_U
    =
    \left.
    \frac{
        \partial \mathcal{U}(\epsilon)
    }{
        \partial \epsilon
    }
    \right|_{\epsilon=0}.
\label{eq:app-gen-jac}
\end{equation}
Here, $J_F$ is a total derivative: it includes both the direct dependence of the generative field on the perturbed observation and the dependence induced through the intermediate generative input $z_s(\epsilon)$.

For sufficiently smooth $\mathcal{F}_\theta$ and $\mathcal{U}$, first-order Taylor expansions give
\begin{equation}
    \mathcal{F}_\theta(\epsilon)
    =
    \mathcal{F}_\theta(0)
    +
    J_F\epsilon
    +
    O(\|\epsilon\|^2)
\end{equation}
and
\begin{equation}
    \mathcal{U}(\epsilon)
    =
    \mathcal{U}(0)
    +
    J_U\epsilon
    +
    O(\|\epsilon\|^2).
\end{equation}
Therefore,
\begin{equation}
\begin{aligned}
\mathcal{F}_\theta(\epsilon)
-
\mathcal{U}(\epsilon)
=
r_g
+
(J_F-J_U)\epsilon
+
O(\|\epsilon\|^2).
\end{aligned}
\end{equation}

For a small zero-mean perturbation and when the generative regression residual $r_g$ is small, taking expectation in \cref{eq:app-gen-loss} yields the local approximation
\begin{equation}
\begin{aligned}
\mathbb{E}_{\epsilon}
\left[
    \mathcal{L}_{\mathrm{aug}}^{\mathrm{gen}}(\epsilon)
\right]
\approx\;&
\mathcal{L}_{\mathrm{aug}}^{\mathrm{gen}}(0)
\\
&+
\frac{1}{2}
\mathbb{E}_{s,\omega}
\left[
    \operatorname{tr}
    \left(
        (J_F-J_U)
        \Sigma_\epsilon
        (J_F-J_U)^\top
    \right)
\right].
\end{aligned}
\label{eq:app-gen-sensitivity}
\end{equation}

Unlike direct trajectory regression, the induced regularizer therefore does not generally penalize $J_F$ alone. Instead, it penalizes the difference between the perturbation response of the learned generative field and that of the surrogate generative target.

The selective structure of PARS appears through the construction of the surrogate trajectory. The suffix of $\hat{\mathbf{a}}_t(\epsilon)$ is encouraged to remain consistent with the demonstrated future, whereas its prefix is allowed to vary in response to the perturbed observation. Under a locally suffix-preserving surrogate construction and the usual generative corruption or interpolation process, the target associated with the suffix is approximately insensitive to the perturbation:
\begin{equation}
    J_{U,\mathrm{suf}}
    \approx
    0.
\label{eq:app-gen-target-suffix}
\end{equation}
In this regime, the suffix component of \cref{eq:app-gen-sensitivity} approximately reduces to
\begin{equation}
\frac{1}{2}
\mathbb{E}_{s,\omega}
\left[
    \operatorname{tr}
    \left(
        J_{F,\mathrm{suf}}
        \Sigma_\epsilon
        J_{F,\mathrm{suf}}^\top
    \right)
\right].
\label{eq:app-gen-suffix-sensitivity}
\end{equation}
Thus, the generative training objective locally penalizes sensitivity associated with the supervised suffix.

In contrast, the surrogate prefix is allowed to vary with the perturbation. Consequently, the intermediate generative input $z_s(\epsilon)$ and, depending on the particular parameterization, the regression target $u_s(\epsilon)$ may also vary with $\epsilon$. The prefix contribution in \cref{eq:app-gen-sensitivity} therefore penalizes the total perturbation-response mismatch $J_{F,\mathrm{pre}}-J_{U,\mathrm{pre}}$. Because $J_F$ includes both the direct dependence on the perturbed observation and the dependence induced through $z_s(\epsilon)$, this term should not in general be interpreted as forcing the direct observation-conditioning sensitivity of the prefix toward zero. This distinction is consistent with the design of PARS: the recovery prefix is allowed to respond to the perturbed state, while the later portion of the trajectory is encouraged to return toward the demonstrated future.

As with SFP, this field-level regularization should not be interpreted as direct regularization of the complete Jacobian of the final generated trajectory. For flow-matching models, the final trajectory is obtained by integrating the learned velocity field, while for diffusion policies it is obtained through a sequence of denoising transitions. Under local regularity of the corresponding generation process, perturbations of the learned field induce locally bounded perturbations of the generated sample. The field-level regularization in \cref{eq:app-gen-suffix-sensitivity} can therefore be viewed as an indirect mechanism for controlling one source of suffix sensitivity in the final generated trajectory.

The exact forms of $z_s$, $u_s$, and $J_U$ depend on the particular flow-matching or diffusion parameterization. Accordingly, \cref{eq:app-gen-sensitivity} is intended as a generic local characterization of surrogate-target generative training rather than as a parameterization-specific identity.

\section{Local Analysis of the Free Prefix}
\label{app:prefix_analysis}
We provide the detailed derivations of the local-consistency bound in \cref{eq:consistency_bound} and the trajectory-planning feasibility condition in \cref{eq:feasibility_condition}.

\paragraph{Local consistency with demonstrated motion.}
Let $\mathbf{q}^{\mathrm{pre}}_\theta(o)=(q_{\theta,0}(o),\ldots,q_{\theta,H_r-1}(o))$ denote the desired-configuration prefix generated by the policy. We define the trajectory-level metric
\begin{equation}
d_{\mathrm{traj}}
\left(
    \mathbf{q},
    \mathbf{q}'
\right)
=
\max_{0\le k < H_r}
\left\|
    q_k-q'_k
\right\|_2 .
\label{eq:app_traj_metric}
\end{equation}
Let $\mathcal{S}^{\mathrm{pre}}_{\mathrm{data}}$ denote the set of configuration prefixes represented in the demonstrations, and define the distance to this set as
\begin{equation}
\operatorname{dist}
\left(
    \mathbf{q},
    \mathcal{S}^{\mathrm{pre}}_{\mathrm{data}}
\right)
=
\inf_{\boldsymbol{\xi}
\in\mathcal{S}^{\mathrm{pre}}_{\mathrm{data}}}
d_{\mathrm{traj}}
\left(
    \mathbf{q},
    \boldsymbol{\xi}
\right).
\end{equation}

For any $\boldsymbol{\xi}\in\mathcal{S}^{\mathrm{pre}}_{\mathrm{data}}$, the triangle inequality gives
\begin{equation}
\begin{aligned}
d_{\mathrm{traj}}
\left(
    \mathbf{q}^{\mathrm{pre}}_\theta(\tilde{o}_t),
    \boldsymbol{\xi}
\right)
\le\;&
d_{\mathrm{traj}}
\left(
    \mathbf{q}^{\mathrm{pre}}_\theta(\tilde{o}_t),
    \mathbf{q}^{\mathrm{pre}}_\theta(o_t)
\right)
\\
&+
d_{\mathrm{traj}}
\left(
    \mathbf{q}^{\mathrm{pre}}_\theta(o_t),
    \boldsymbol{\xi}
\right).
\end{aligned}
\end{equation}
Taking the infimum over $\boldsymbol{\xi}\in\mathcal{S}^{\mathrm{pre}}_{\mathrm{data}}$ and applying the assumptions in \cref{eq:prefix_assumptions} yields
\begin{equation}
\begin{aligned}
\operatorname{dist}
\left(
    \mathbf{q}^{\mathrm{pre}}_\theta(\tilde{o}_t),
    \mathcal{S}^{\mathrm{pre}}_{\mathrm{data}}
\right)
\le\;&
d_{\mathrm{traj}}
\left(
    \mathbf{q}^{\mathrm{pre}}_\theta(\tilde{o}_t),
    \mathbf{q}^{\mathrm{pre}}_\theta(o_t)
\right)
\\
&+
\operatorname{dist}
\left(
    \mathbf{q}^{\mathrm{pre}}_\theta(o_t),
    \mathcal{S}^{\mathrm{pre}}_{\mathrm{data}}
\right)
\\
\le\;&
\eta + L_q\|\epsilon\|_2 .
\end{aligned}
\label{eq:app_consistency}
\end{equation}
Since each perturbed coordinate satisfies $|\epsilon_i|\le\delta_e$ and the perturbation has dimension $d_e$,
\begin{equation}
\|\epsilon\|_2
\le
\sqrt{d_e}\,\delta_e .
\end{equation}
Substituting this bound into \cref{eq:app_consistency} gives
\begin{equation}
\operatorname{dist}
\left(
    \mathbf{q}^{\mathrm{pre}}_\theta(\tilde{o}_t),
    \mathcal{S}^{\mathrm{pre}}_{\mathrm{data}}
\right)
\le
\eta + L_q\sqrt{d_e}\,\delta_e ,
\end{equation}
which recovers \cref{eq:consistency_bound}.

\paragraph{Feasibility under trajectory-planning limits.}
We next show how the same local-sensitivity assumption bounds changes in the joint-space speed and acceleration of the free prefix. For a fixed control interval $\Delta t$, define
\begin{equation}
v_{\theta,k}(o)
=
\frac{
    q_{\theta,k+1}(o)-q_{\theta,k}(o)
}{
    \Delta t
},
\qquad
0\le k < H_r-1,
\label{eq:app_velocity_def}
\end{equation}
and
\begin{equation}
a_{\theta,k}(o)
=
\frac{
    q_{\theta,k+2}(o)
    -2q_{\theta,k+1}(o)
    +q_{\theta,k}(o)
}{
    \Delta t^2
},
\qquad
0\le k < H_r-2.
\label{eq:app_acceleration_def}
\end{equation}

For convenience, define the perturbation-induced change in the predicted configuration at step $k$ as
\begin{equation}
\Delta q_{\theta,k}
=
q_{\theta,k}(\tilde{o}_t)
-
q_{\theta,k}(o_t).
\end{equation}
From \cref{eq:prefix_assumptions,eq:app_traj_metric},
\begin{equation}
\left\|
    \Delta q_{\theta,k}
\right\|_2
\le
L_q\|\epsilon\|_2
\end{equation}
for every step in the free prefix.

The corresponding change in velocity is
\begin{equation}
v_{\theta,k}(\tilde{o}_t)
-
v_{\theta,k}(o_t)
=
\frac{
    \Delta q_{\theta,k+1}
    -
    \Delta q_{\theta,k}
}{
    \Delta t
}.
\end{equation}
Therefore,
\begin{align}
\left\|
    v_{\theta,k}(\tilde{o}_t)
    -
    v_{\theta,k}(o_t)
\right\|_2
&\le
\frac{
    \|\Delta q_{\theta,k+1}\|_2
    +
    \|\Delta q_{\theta,k}\|_2
}{
    \Delta t
}
\nonumber\\
&\le
\frac{
    2L_q\|\epsilon\|_2
}{
    \Delta t
}
\nonumber\\
&\le
\frac{
    2L_q\sqrt{d_e}\,\delta_e
}{
    \Delta t
}.
\label{eq:app_velocity_bound}
\end{align}

Similarly,
\begin{equation}
a_{\theta,k}(\tilde{o}_t)
-
a_{\theta,k}(o_t)
=
\frac{
    \Delta q_{\theta,k+2}
    -
    2\Delta q_{\theta,k+1}
    +
    \Delta q_{\theta,k}
}{
    \Delta t^2
},
\end{equation}
which gives
\begin{align}
\left\|
    a_{\theta,k}(\tilde{o}_t)
    -
    a_{\theta,k}(o_t)
\right\|_2
&\le
\frac{
    \|\Delta q_{\theta,k+2}\|_2
    +
    2\|\Delta q_{\theta,k+1}\|_2
    +
    \|\Delta q_{\theta,k}\|_2
}{
    \Delta t^2
}
\nonumber\\
&\le
\frac{
    4L_q\|\epsilon\|_2
}{
    \Delta t^2
}
\nonumber\\
&\le
\frac{
    4L_q\sqrt{d_e}\,\delta_e
}{
    \Delta t^2
}.
\label{eq:app_acceleration_bound}
\end{align}

Let $v^{\max}$ and $a^{\max}$ denote the joint-space speed and acceleration limits used for trajectory planning. As stated in the main text, suppose that the nominal prefix satisfies these limits with positive margins $m_v$ and $m_a$, such that
\begin{equation}
\left\|
    v_{\theta,k}(o_t)
\right\|_2
\le
v^{\max}-m_v,
\qquad
\left\|
    a_{\theta,k}(o_t)
\right\|_2
\le
a^{\max}-m_a
\label{eq:app_nominal_margin}
\end{equation}
for all valid steps $k$.

Using \cref{eq:app_velocity_bound},
\begin{align}
\left\|
    v_{\theta,k}(\tilde{o}_t)
\right\|_2
&\le
\left\|
    v_{\theta,k}(o_t)
\right\|_2
+
\left\|
    v_{\theta,k}(\tilde{o}_t)
    -
    v_{\theta,k}(o_t)
\right\|_2
\nonumber\\
&\le
v^{\max}
-
m_v
+
\frac{
    2L_q\sqrt{d_e}\,\delta_e
}{
    \Delta t
}.
\end{align}
Hence,
\begin{equation}
\frac{
    2L_q\sqrt{d_e}\,\delta_e
}{
    \Delta t
}
\le
m_v
\end{equation}
is sufficient to ensure
\begin{equation}
\left\|
    v_{\theta,k}(\tilde{o}_t)
\right\|_2
\le
v^{\max}.
\end{equation}

Likewise, from \cref{eq:app_acceleration_bound},
\begin{align}
\left\|
    a_{\theta,k}(\tilde{o}_t)
\right\|_2
&\le
\left\|
    a_{\theta,k}(o_t)
\right\|_2
+
\left\|
    a_{\theta,k}(\tilde{o}_t)
    -
    a_{\theta,k}(o_t)
\right\|_2
\nonumber\\
&\le
a^{\max}
-
m_a
+
\frac{
    4L_q\sqrt{d_e}\,\delta_e
}{
    \Delta t^2
}.
\end{align}
Thus,
\begin{equation}
\frac{
    4L_q\sqrt{d_e}\,\delta_e
}{
    \Delta t^2
}
\le
m_a
\end{equation}
is sufficient to ensure
\begin{equation}
\left\|
    a_{\theta,k}(\tilde{o}_t)
\right\|_2
\le
a^{\max}.
\end{equation}
Together, these conditions recover \cref{eq:feasibility_condition}.

\paragraph{Interpretation of the assumptions.}
The analysis is conditional on three local properties. First, the nominal BC prediction is assumed to remain close to configuration trajectories represented in the demonstrations. Second, a small proprioceptive perturbation is assumed to induce only a proportionally small change in the predicted configuration prefix. Third, the nominal prefix is assumed to satisfy the joint-space speed and acceleration limits used for trajectory planning with positive margins. These assumptions are not guaranteed by the suffix-anchoring objective itself.

\paragraph{Scope of the analysis.}
The two results characterize complementary properties of the free recovery prefix. The local-consistency bound limits how far the perturbed prefix can move from configuration trajectories represented in the demonstrations, whereas the feasibility condition ensures that its joint-space speed and acceleration remain within the specified trajectory-planning limits when the perturbation is sufficiently small. Neither result requires the free prefix to reproduce a particular demonstrated trajectory, leaving room for corrective motion. However, the analysis is local and does not guarantee task success, collision avoidance, or full dynamic feasibility under arbitrarily large perturbations. In particular, satisfying these joint-space speed and acceleration limits alone does not guarantee perfect tracking by the low-level controller.

\section{Experimental Settings}

\subsection{Training and Policy Settings}
\label{app:training-details}

\begin{table}[H]
\centering
\caption{Common policy-training and success-rate evaluation settings.}
\label{tab:training_setup}
\begin{tabular}{ll}
\hline
\textbf{Setting} & \textbf{Value} \\
\hline
Training steps & $40{,}000$ gradient steps \\
Optimizer & AdamW \\
Learning rate & $3\times10^{-4}$ \\
Learning-rate schedule & Cosine decay to $10^{-6}$ \\
Batch size & $128$ \\
Random seed & $0$ \\
Policy training & Separate policy trained for each task \\
Evaluation episodes & $50$ per task \\
Maximum episode length & $300$ environment steps \\
\hline
\end{tabular}
\end{table}

\begin{table}[H]
\centering
\caption{
Policy-specific settings. Architecture and policy hyperparameters are selected
for the corresponding unaugmented baseline and held fixed across the
unaugmented, PARS, and noise-augmentation conditions.
}
\label{tab:hyper}
\begin{tabular}{@{}l p{0.25\linewidth} p{0.25\linewidth} p{0.31\linewidth}@{}}
\hline
\textbf{Method}
& \textbf{Architecture}
& \textbf{Generation / inference}
& \textbf{Method-specific settings} \\
\hline

FP
& Width 144, 4 layers, 8 heads
& 10 ODE steps
& --- \\

SFP
& MLP field, width 256
& 3 Euler substeps per action
& $\sigma_0=\sigma_1=0.1$ \\

TP
& Width 152, 4 layers, 8 heads
& ---
& --- \\

ACT
& Width 256, 4 layers, 8 heads
& ---
& CVAE latent dim.\ 32, KL weight 10 \\

DP
& U-Net $(128,256,512)$; per-task encoder width (256 default)
& 100 training / 20 inference steps
& Squared-cosine noise schedule \\

KDPE
& DP checkpoint
& 32 candidate chunks
& KDE bandwidth 0.2; $20\times$ bandwidth for gripper dimension \\
\hline
\end{tabular}
\end{table}

For DP, we use the squared-cosine noise schedule rather than the linear
schedule because the latter is calibrated for $T=1000$ and leaves
$\bar{\alpha}$ far from zero with $100$ diffusion steps, causing sampling
to start outside the noise range encountered during training.

\subsection{PARS and Augmentation Settings}
\label{app:augmentation-settings}

Table~\ref{tab:pars_setup} summarizes the PARS and augmentation settings used in our experiments. For each perturbed sample, the perturbation scale $\delta_e$ is sampled from the task-specific range $[\sigma_{\mathrm{lo}},\sigma_{\mathrm{hi}}]$, after which the proprioceptive perturbation is drawn as $\epsilon\sim\mathcal{U}(-\delta_e,\delta_e)$. Each training sample uses the augmented objective with task-specific probability $p$ and the standard BC objective otherwise. The expected objective is therefore $(1-p)\mathcal{L}_{\mathrm{BC}} +p\mathcal{L}_{\mathrm{aug}}$; after factoring out $(1-p)$, the relative augmented-loss coefficient in Eq.~\ref{eq:pars} is $\lambda_{\mathrm{aug}}=p/(1-p)$. For each task, the augmentation setting is determined by grid search. The selected PARS settings used for Table~\ref{tab:main} are reported in Tables~\ref{tab:best_settings_1} and~\ref{tab:best_settings_2}.

\begin{table}[H]
\centering
\caption{PARS and augmentation settings. Settings marked ``per task'' are selected separately for each task.}
\label{tab:pars_setup}
\begin{tabular}{@{}l l l@{}}
\hline
\textbf{Setting} & \textbf{Applies to} & \textbf{Value} \\
\hline

Prediction horizon $H$
& All methods
& $20$ \\

Free recovery prefix $H_r$
& PARS
& $10$ \\

& Noise augmentation
& $0$ \\

\hline

Perturbation probability $p$
& PARS, noise augmentation
& Per task \\

Perturbation scale range
& PARS, noise augmentation
& Per task \\

\hline

Inner refinement steps
& FP, DP
& Per task \\

Inner refinement step size
& FP, DP
& $0.1$ \\

Augmentation warm-up
& DP
& Per task \\

\hline
\end{tabular}
\end{table}

DP uses an initial warm-up period before augmentation is enabled, as applying
refinement to samples from the poorly trained policy early in training was
empirically less stable in preliminary runs. For SFP, the stepwise integration
required to obtain the state at $\tau_r$ is performed on $16$ perturbed
samples per batch because sequential integration is computationally more
expensive.

\subsection{Recovery and Free-Prefix Analysis}
\label{app:prefix-kinematics}

Figures~\ref{fig:recovery} and~\ref{fig:prefix_kinematics} analyze the predicted action sequences at off-nominal observations reached during closed-loop execution. Both analyses use the predicted action sequence rather than the subsequently executed rollout, since the quantities of interest describe the motion proposed by the policy from an off-nominal observation.

\paragraph{Off-nominal observation collection.}
The off-nominal observations are not synthetically generated. For each trainable policy class, we collect observations from $20$ rollouts per task of the corresponding unaugmented policy. The configurations visited at action-sequence boundaries are ranked by their nearest-neighbor distance to the demonstrated joint configurations in $z$-scored joint space. We retain the top quartile and query all compared variants at the same recorded observations. The comparison is therefore paired at the level of individual observations, and the analyzed states reflect deviations actually reached during closed-loop execution. For KDPE, the observations corresponding to DP are used, since KDPE shares the DP checkpoint and differs only in its inference-time selection rule.

\paragraph{Kinematic metrics and reference limits.}
We use a control period of $\Delta t=0.05\,\mathrm{s}$. The RLBench demonstration planner constrains the rate of progress along the planned joint-space path rather than the velocity of each individual joint. We therefore measure the Euclidean norm over the seven arm joints, $\lVert\Delta q\rVert_2/\Delta t$ for speed and $\lVert\Delta^2 q\rVert_2/\Delta t^2$ for acceleration. Using the maximum individual-joint value would instead compare against a constraint that was not used to generate the demonstrations.

The dashed reference lines in Figure~\ref{fig:prefix_kinematics} correspond to the trajectory-planning limits used by the RLBench demonstration generator: $1.0\,\mathrm{rad/s}$ for joint-space speed and $4.0\,\mathrm{rad/s^2}$ for joint-space acceleration. These values are planning constraints on the demonstration trajectories rather than hardware limits of the robot. Numerical differentiation at $20\,\mathrm{Hz}$ produces small apparent exceedances around the speed limit: $22\%$ of demonstrated steps lie slightly above $1.0\,\mathrm{rad/s}$, whereas only $2.7\%$ exceed $4.0\,\mathrm{rad/s^2}$ in acceleration. We therefore separate the speed range immediately above the planning limit from the more aggressive region above $1.25\,\mathrm{rad/s}$.

\paragraph{Recovery-distance normalization.}
For each predicted step $k$, we compute the predicted joint configuration as $\hat q_k=q_{\mathrm{init}}+\sum_{i=0}^{k}\Delta q_i$, where $q_{\mathrm{init}}$ is the joint configuration at the observed action-sequence boundary, and measure its nearest-neighbor distance to the demonstrated joint configurations. The distance is divided by its value at $k=0$ to make the curves comparable across tasks with different absolute distance scales. Consequently, a normalized value of $1.0$ corresponds to the reference distance at $k=0$, values below $1.0$ indicate movement toward the demonstrated configurations, and values above $1.0$ indicate further deviation. Values are first averaged across observations within each task and then across the 51 tasks.

\subsection{Selected Augmentation Settings for Each Task}
\label{app:best-settings}
Tables~\ref{tab:best_settings_1} and~\ref{tab:best_settings_2} report the task-specific augmentation settings used for the PARS results in Table~\ref{tab:main}. DP and KDPE share the same settings because KDPE is applied only at inference time to the corresponding DP checkpoint.

\begin{table}[H]
\centering
\caption{Task-specific PARS settings for FP, SFP, and TP. Range denotes $\sigma_{\mathrm{lo}}$--$\sigma_{\mathrm{hi}}$, $p$ is the perturbation probability, and $K$ is the number of inner refinement steps.}
\label{tab:best_settings_1}
\setlength{\tabcolsep}{3pt}
\begin{tabular}{@{}lccccccc@{}}
\hline
\multirow{2}{*}{\textbf{RLBench Task}}
& \multicolumn{3}{c}{\textbf{FP}}
& \multicolumn{2}{c}{\textbf{SFP}}
& \multicolumn{2}{c}{\textbf{TP}} \\
\cline{2-4}\cline{5-6}\cline{7-8}
& Range & $p$ & $K$ & Range & $p$ & Range & $p$ \\
\hline
basketball in hoop & 0.05--0.15 & 0.65 & 5 & 0.03--0.08 & 0.65 & 0.04--0.12 & 0.5 \\
beat the buzz & 0.02--0.06 & 0.35 & 5 & 0.03--0.08 & 0.35 & 0.01--0.04 & 0.15 \\
change clock & 0.03--0.1 & 0.5 & 25 & 0.05--0.08 & 0.65 & 0.04--0.12 & 0.65 \\
close box & 0.03--0.1 & 0.5 & 25 & 0.05--0.08 & 0.65 & 0.02--0.06 & 0.5 \\
close door & 0.02--0.06 & 0.5 & 5 & 0.05--0.15 & 0.3 & 0.05--0.15 & 0.5 \\
close drawer & 0.03--0.1 & 0.5 & 25 & 0.04--0.12 & 0.5 & 0.03--0.1 & 0.5 \\
close fridge & 0.02--0.06 & 0.35 & 5 & 0.05--0.15 & 0.65 & 0.02--0.04 & 0.35 \\
close grill & 0.05--0.08 & 0.5 & 40 & 0.04--0.12 & 0.5 & 0.03--0.1 & 0.5 \\
close jar & 0.03--0.1 & 0.5 & 5 & 0.03--0.08 & 0.5 & 0.03--0.1 & 0.5 \\
close laptop lid & 0.03--0.08 & 0.5 & 25 & 0.03--0.1 & 0.3 & 0.02--0.04 & 0.35 \\
close microwave & 0.03--0.08 & 0.5 & 25 & 0.03--0.08 & 0.5 & 0.02--0.06 & 0.35 \\
lamp off & 0.03--0.1 & 0.5 & 15 & 0.03--0.1 & 0.65 & 0.05--0.15 & 0.35 \\
lamp on & 0.03--0.1 & 0.5 & 15 & 0.05--0.15 & 0.3 & 0.05--0.15 & 0.5 \\
lift numbered block & 0.03--0.1 & 0.5 & 25 & 0.03--0.1 & 0.65 & 0.05--0.15 & 0.5 \\
meat off grill & 0.03--0.1 & 0.5 & 25 & 0.03--0.08 & 0.5 & 0.02--0.04 & 0.15 \\
meat on grill & 0.01--0.04 & 0.5 & 15 & 0.01--0.03 & 0.3 & 0.02--0.06 & 0.15 \\
open box & 0.02--0.06 & 0.35 & 5 & 0.03--0.1 & 0.5 & 0.03--0.1 & 0.25 \\
open door & 0.03--0.1 & 0.5 & 5 & 0.05--0.15 & 0.5 & 0.02--0.06 & 0.35 \\
open drawer & 0.05--0.15 & 0.5 & 25 & 0.03--0.1 & 0.3 & 0.05--0.08 & 0.65 \\
open fridge & 0.03--0.1 & 0.5 & 15 & 0.05--0.08 & 0.65 & 0.03--0.08 & 0.35 \\
open grill & 0.04--0.12 & 0.5 & 25 & 0.04--0.12 & 0.5 & 0.01--0.04 & 0.5 \\
open jar & 0.02--0.06 & 0.35 & 5 & 0.03--0.08 & 0.65 & 0.03--0.1 & 0.5 \\
open microwave & 0.03--0.1 & 0.5 & 25 & 0.03--0.08 & 0.15 & 0.05--0.08 & 0.5 \\
open oven & 0.01--0.04 & 0.5 & 15 & 0.03--0.08 & 0.5 & 0.02--0.06 & 0.35 \\
open washing machine & 0.03--0.1 & 0.5 & 25 & 0.03--0.1 & 0.65 & 0.01--0.04 & 0.25 \\
open wine bottle & 0.03--0.08 & 0.5 & 25 & 0.03--0.08 & 0.65 & 0.01--0.04 & 0.25 \\
phone on base & 0.03--0.1 & 0.5 & 25 & 0.04--0.12 & 0.35 & 0.02--0.06 & 0.15 \\
pick and lift & 0.03--0.1 & 0.5 & 15 & 0.03--0.08 & 0.65 & 0.03--0.08 & 0.35 \\
pick and lift small & 0.03--0.1 & 0.5 & 5 & 0.05--0.08 & 0.5 & 0.04--0.12 & 0.65 \\
pick up cup & 0.03--0.1 & 0.5 & 25 & 0.05--0.08 & 0.5 & 0.03--0.08 & 0.5 \\
press switch & 0.03--0.1 & 0.5 & 15 & 0.04--0.12 & 0.5 & 0.03--0.08 & 0.5 \\
push button & 0.03--0.1 & 0.5 & 5 & 0.03--0.08 & 0.65 & 0.05--0.08 & 0.65 \\
push buttons & 0.03--0.1 & 0.5 & 5 & 0.01--0.03 & 0.3 & 0.01--0.04 & 0.25 \\
put item in drawer & 0.04--0.12 & 0.5 & 15 & 0.03--0.1 & 0.65 & 0.02--0.06 & 0.5 \\
put knife on chopping board & 0.03--0.1 & 0.5 & 25 & 0.05--0.08 & 0.65 & 0.02--0.04 & 0.25 \\
put money in safe & 0.03--0.1 & 0.5 & 25 & 0.04--0.12 & 0.5 & 0.03--0.1 & 0.15 \\
put rubbish in bin & 0.03--0.1 & 0.5 & 15 & 0.04--0.12 & 0.5 & 0.02--0.06 & 0.15 \\
put umbrella in umbrella stand & 0.01--0.04 & 0.5 & 15 & 0.04--0.12 & 0.5 & 0.03--0.1 & 0.25 \\
reach and drag & 0.03--0.1 & 0.5 & 40 & 0.01--0.03 & 0.3 & 0.04--0.12 & 0.5 \\
reach target & 0.02--0.06 & 0.35 & 5 & 0.03--0.08 & 0.5 & 0.03--0.1 & 0.5 \\
slide block to target & 0.03--0.1 & 0.5 & 5 & 0.03--0.08 & 0.5 & 0.04--0.12 & 0.5 \\
stack blocks & 0.03--0.1 & 0.5 & 5 & 0.03--0.08 & 0.5 & 0.03--0.1 & 0.5 \\
take item out of drawer & 0.03--0.1 & 0.65 & 25 & 0.05--0.08 & 0.5 & 0.02--0.06 & 0.35 \\
take lid off saucepan & 0.05--0.15 & 0.65 & 5 & 0.04--0.12 & 0.5 & 0.03--0.1 & 0.5 \\
take umbrella out of umbrella stand & 0.03--0.1 & 0.5 & 15 & 0.05--0.08 & 0.5 & 0.03--0.1 & 0.5 \\
toilet seat down & 0.02--0.06 & 0.35 & 5 & 0.03--0.08 & 0.65 & 0.03--0.08 & 0.35 \\
toilet seat up & 0.03--0.1 & 0.5 & 15 & 0.03--0.08 & 0.65 & 0.02--0.06 & 0.5 \\
turn oven on & 0.03--0.1 & 0.5 & 5 & 0.03--0.08 & 0.5 & 0.04--0.12 & 0.5 \\
turn tap & 0.05--0.15 & 0.65 & 5 & 0.03--0.1 & 0.65 & 0.05--0.15 & 0.5 \\
unplug charger & 0.03--0.1 & 0.5 & 5 & 0.04--0.12 & 0.65 & 0.05--0.08 & 0.65 \\
water plants & 0.03--0.1 & 0.5 & 25 & 0.03--0.1 & 0.65 & 0.05--0.08 & 0.65 \\
\hline

\end{tabular}
\end{table}

\begin{table}[H]
\centering
\caption{Task-specific PARS settings for ACT and DP/KDPE. Range denotes $\sigma_{\mathrm{lo}}$--$\sigma_{\mathrm{hi}}$; $K$ is the number of inner refinement steps. Warm-up is reported in gradient steps, with ``k'' denoting $1{,}000$.}
\label{tab:best_settings_2}
\setlength{\tabcolsep}{3pt}
\begin{tabular}{@{}lcccccc@{}}
\hline
\multirow{2}{*}{\textbf{RLBench Task}}
& \multicolumn{2}{c}{\textbf{ACT}}
& \multicolumn{4}{c}{\textbf{DP / KDPE}} \\
\cline{2-3}\cline{4-7}
& Range & $p$ & Range & $p$ & $K$ & Warm-up \\
\hline
basketball in hoop & 0.02--0.06 & 0.35 & 0.03--0.1 & 0.5 & 25 & 15k \\
beat the buzz & 0.03--0.08 & 0.1 & 0.05--0.08 & 0.5 & 5 & 15k \\
change clock & 0.01--0.04 & 0.25 & 0.03--0.1 & 0.5 & 15 & 15k \\
close box & 0.02--0.04 & 0.25 & 0.03--0.1 & 0.5 & 25 & 15k \\
close door & 0.02--0.04 & 0.35 & 0.05--0.15 & 0.5 & 15 & 15k \\
close drawer & 0.05--0.15 & 0.65 & 0.03--0.1 & 0.5 & 25 & 15k \\
close fridge & 0.03--0.1 & 0.5 & 0.03--0.1 & 0.65 & 25 & 20k \\
close grill & 0.02--0.04 & 0.5 & 0.04--0.12 & 0.5 & 25 & 15k \\
close jar & 0.03--0.08 & 0.5 & 0.03--0.1 & 0.5 & 25 & 15k \\
close laptop lid & 0.03--0.1 & 0.65 & 0.03--0.1 & 0.5 & 25 & 20k \\
close microwave & 0.01--0.04 & 0.15 & 0.03--0.1 & 0.65 & 5 & 15k \\
lamp off & 0.01--0.04 & 0.5 & 0.03--0.08 & 0.5 & 5 & 15k \\
lamp on & 0.04--0.12 & 0.15 & 0.03--0.1 & 0.5 & 15 & 15k \\
lift numbered block & 0.02--0.04 & 0.15 & 0.03--0.08 & 0.65 & 25 & 15k \\
meat off grill & 0.01--0.03 & 0.25 & 0.05--0.08 & 0.65 & 25 & 15k \\
meat on grill & 0.01--0.03 & 0.25 & 0.03--0.08 & 0.5 & 25 & 15k \\
open box & 0.02--0.04 & 0.35 & 0.03--0.08 & 0.5 & 25 & 15k \\
open door & 0.01--0.04 & 0.25 & 0.03--0.08 & 0.5 & 25 & 15k \\
open drawer & 0.02--0.04 & 0.05 & 0.04--0.12 & 0.5 & 15 & 20k \\
open fridge & 0.03--0.08 & 0.65 & 0.04--0.12 & 0.5 & 25 & 15k \\
open grill & 0.02--0.04 & 0.25 & 0.03--0.08 & 0.5 & 15 & 20k \\
open jar & 0.05--0.08 & 0.5 & 0.03--0.1 & 0.5 & 25 & 15k \\
open microwave & 0.02--0.06 & 0.1 & 0.04--0.12 & 0.5 & 25 & 15k \\
open oven & 0.02--0.06 & 0.65 & 0.05--0.08 & 0.65 & 25 & 15k \\
open washing machine & 0.03--0.1 & 0.5 & 0.03--0.1 & 0.5 & 25 & 15k \\
open wine bottle & 0.03--0.08 & 0.25 & 0.03--0.08 & 0.5 & 5 & 20k \\
phone on base & 0.03--0.1 & 0.15 & 0.04--0.12 & 0.5 & 25 & 15k \\
pick and lift & 0.05--0.08 & 0.35 & 0.04--0.12 & 0.5 & 15 & 15k \\
pick and lift small & 0.03--0.08 & 0.5 & 0.03--0.1 & 0.5 & 25 & 15k \\
pick up cup & 0.03--0.08 & 0.35 & 0.03--0.08 & 0.5 & 5 & 15k \\
press switch & 0.02--0.06 & 0.35 & 0.04--0.12 & 0.5 & 15 & 20k \\
push button & 0.05--0.08 & 0.5 & 0.03--0.1 & 0.5 & 25 & 20k \\
push buttons & 0.03--0.08 & 0.2 & 0.03--0.1 & 0.5 & 25 & 20k \\
put item in drawer & 0.03--0.1 & 0.5 & 0.03--0.08 & 0.5 & 5 & 15k \\
put knife on chopping board & 0.04--0.12 & 0.25 & 0.05--0.08 & 0.65 & 15 & 15k \\
put money in safe & 0.01--0.03 & 0.15 & 0.04--0.12 & 0.5 & 15 & 15k \\
put rubbish in bin & 0.03--0.1 & 0.15 & 0.04--0.12 & 0.5 & 15 & 20k \\
put umbrella in umbrella stand & 0.03--0.08 & 0.35 & 0.04--0.12 & 0.5 & 15 & 20k \\
reach and drag & 0.01--0.04 & 0.35 & 0.03--0.1 & 0.5 & 25 & 15k \\
reach target & 0.03--0.1 & 0.5 & 0.03--0.1 & 0.5 & 25 & 15k \\
slide block to target & 0.05--0.15 & 0.5 & 0.03--0.08 & 0.5 & 15 & 15k \\
stack blocks & 0.03--0.1 & 0.5 & 0.03--0.1 & 0.5 & 25 & 15k \\
take item out of drawer & 0.03--0.1 & 0.25 & 0.04--0.12 & 0.5 & 15 & 20k \\
take lid off saucepan & 0.02--0.04 & 0.25 & 0.03--0.08 & 0.5 & 5 & 15k \\
take umbrella out of umbrella stand & 0.02--0.06 & 0.5 & 0.03--0.1 & 0.5 & 15 & 15k \\
toilet seat down & 0.01--0.04 & 0.35 & 0.03--0.1 & 0.65 & 5 & 20k \\
toilet seat up & 0.03--0.1 & 0.25 & 0.03--0.1 & 0.5 & 5 & 15k \\
turn oven on & 0.01--0.03 & 0.35 & 0.05--0.08 & 0.5 & 5 & 15k \\
turn tap & 0.05--0.08 & 0.65 & 0.03--0.1 & 0.5 & 25 & 20k \\
unplug charger & 0.02--0.06 & 0.65 & 0.04--0.12 & 0.5 & 15 & 20k \\
water plants & 0.01--0.04 & 0.25 & 0.03--0.1 & 0.5 & 25 & 15k \\
\hline
\end{tabular}
\end{table}

\section{Full results}
\begin{table}[H]
\centering
\setlength{\tabcolsep}{1pt}
\caption{Per-task success rate (\%) on all 51 RLBench tasks.}
\label{tab:pertask}
\begin{tabular}{l cc cc cc}
\hline
\multirow{2}{*}{RLBench Task}
& \multicolumn{2}{c}{Flow-based}
& \multicolumn{2}{c}{Transformer-based}
& \multicolumn{2}{c}{Diffusion-based} \\
\cline{2-3}\cline{4-5}\cline{6-7}
& FP & SFP & TP & ACT & DP & KDPE \\
\hline
basketball in hoop & 94$\to$\textbf{96} & 96$\to$\textbf{96} & 94$\to$\textbf{96} & 96$\to$\textbf{96} & 94$\to$84 & 96$\to$\textbf{98} \\
beat the buzz & 66$\to$\textbf{74} & 48$\to$\textbf{58} & 76$\to$72 & 78$\to$\textbf{90} & 28$\to$\textbf{52} & 38$\to$\textbf{50} \\
change clock & 6$\to$\textbf{30} & 8$\to$\textbf{22} & 28$\to$\textbf{34} & 18$\to$\textbf{22} & 6$\to$\textbf{24} & 20$\to$\textbf{26} \\
close box & 88$\to$\textbf{88} & 78$\to$\textbf{90} & 90$\to$\textbf{96} & 92$\to$\textbf{92} & 68$\to$\textbf{88} & 82$\to$\textbf{94} \\
close door & 12$\to$\textbf{18} & 16$\to$\textbf{16} & 4$\to$\textbf{38} & 12$\to$\textbf{24} & 10$\to$\textbf{14} & 10$\to$4 \\
close drawer & 96$\to$\textbf{100} & 98$\to$\textbf{100} & 100$\to$\textbf{100} & 100$\to$\textbf{100} & 100$\to$\textbf{100} & 100$\to$\textbf{100} \\
close fridge & 88$\to$\textbf{98} & 78$\to$\textbf{96} & 90$\to$\textbf{100} & 94$\to$\textbf{96} & 94$\to$\textbf{94} & 98$\to$92 \\
close grill & 94$\to$92 & 90$\to$\textbf{90} & 90$\to$\textbf{92} & 88$\to$\textbf{98} & 74$\to$\textbf{80} & 74$\to$\textbf{88} \\
close jar & 0$\to$\textbf{0} & 0$\to$\textbf{0} & 0$\to$\textbf{0} & 0$\to$\textbf{0} & 0$\to$\textbf{0} & 0$\to$\textbf{0} \\
close laptop lid & 98$\to$96 & 100$\to$92 & 96$\to$\textbf{96} & 98$\to$\textbf{98} & 76$\to$\textbf{98} & 82$\to$\textbf{100} \\
close microwave & 100$\to$\textbf{100} & 88$\to$\textbf{98} & 96$\to$\textbf{98} & 100$\to$\textbf{100} & 100$\to$94 & 100$\to$\textbf{100} \\
lamp off & 84$\to$\textbf{92} & 72$\to$\textbf{86} & 84$\to$\textbf{90} & 80$\to$\textbf{88} & 88$\to$82 & 86$\to$\textbf{90} \\
lamp on & 80$\to$\textbf{90} & 80$\to$78 & 78$\to$\textbf{86} & 86$\to$82 & 48$\to$\textbf{80} & 56$\to$\textbf{88} \\
lift numbered block & 44$\to$\textbf{60} & 32$\to$\textbf{44} & 22$\to$\textbf{44} & 24$\to$\textbf{52} & 42$\to$36 & 30$\to$\textbf{34} \\
meat off grill & 96$\to$\textbf{98} & 78$\to$\textbf{90} & 94$\to$\textbf{96} & 92$\to$\textbf{100} & 70$\to$\textbf{74} & 90$\to$74 \\
meat on grill & 72$\to$\textbf{90} & 70$\to$\textbf{90} & 78$\to$\textbf{82} & 52$\to$\textbf{92} & 80$\to$\textbf{96} & 80$\to$\textbf{98} \\
open box & 98$\to$\textbf{100} & 86$\to$84 & 90$\to$\textbf{100} & 98$\to$\textbf{100} & 78$\to$\textbf{88} & 90$\to$\textbf{96} \\
open door & 86$\to$\textbf{96} & 82$\to$\textbf{90} & 94$\to$\textbf{100} & 88$\to$\textbf{88} & 74$\to$\textbf{76} & 88$\to$\textbf{92} \\
open drawer & 64$\to$\textbf{66} & 58$\to$\textbf{58} & 54$\to$\textbf{76} & 66$\to$\textbf{68} & 38$\to$\textbf{60} & 48$\to$\textbf{68} \\
open fridge & 10$\to$\textbf{30} & 2$\to$\textbf{18} & 16$\to$\textbf{32} & 14$\to$\textbf{34} & 6$\to$\textbf{10} & 10$\to$4 \\
open grill & 52$\to$\textbf{58} & 46$\to$\textbf{56} & 56$\to$\textbf{74} & 38$\to$\textbf{40} & 12$\to$\textbf{30} & 6$\to$\textbf{18} \\
open jar & 0$\to$\textbf{2} & 0$\to$\textbf{0} & 0$\to$\textbf{0} & 0$\to$\textbf{0} & 0$\to$\textbf{0} & 0$\to$\textbf{0} \\
open microwave & 92$\to$\textbf{92} & 96$\to$\textbf{98} & 92$\to$\textbf{94} & 96$\to$\textbf{96} & 62$\to$\textbf{78} & 80$\to$\textbf{82} \\
open oven & 18$\to$\textbf{30} & 10$\to$\textbf{20} & 18$\to$\textbf{58} & 36$\to$\textbf{68} & 18$\to$\textbf{20} & 22$\to$\textbf{32} \\
open washing machine & 34$\to$\textbf{56} & 18$\to$\textbf{44} & 46$\to$\textbf{64} & 46$\to$\textbf{58} & 4$\to$\textbf{18} & 24$\to$\textbf{38} \\
open wine bottle & 66$\to$\textbf{68} & 38$\to$\textbf{74} & 52$\to$\textbf{82} & 50$\to$\textbf{84} & 52$\to$\textbf{62} & 38$\to$\textbf{66} \\
phone on base & 58$\to$\textbf{66} & 22$\to$20 & 44$\to$20 & 44$\to$\textbf{58} & 30$\to$\textbf{52} & 34$\to$\textbf{44} \\
pick and lift & 66$\to$\textbf{82} & 32$\to$\textbf{86} & 54$\to$\textbf{78} & 44$\to$\textbf{90} & 46$\to$\textbf{58} & 56$\to$\textbf{68} \\
pick and lift small & 0$\to$\textbf{0} & 2$\to$0 & 0$\to$\textbf{2} & 0$\to$\textbf{2} & 0$\to$\textbf{0} & 0$\to$\textbf{0} \\
pick up cup & 90$\to$88 & 74$\to$\textbf{86} & 78$\to$\textbf{88} & 82$\to$\textbf{90} & 82$\to$\textbf{92} & 94$\to$\textbf{94} \\
press switch & 34$\to$\textbf{78} & 18$\to$\textbf{62} & 50$\to$\textbf{82} & 68$\to$\textbf{82} & 38$\to$\textbf{56} & 62$\to$60 \\
push button & 86$\to$\textbf{94} & 82$\to$\textbf{88} & 82$\to$\textbf{98} & 80$\to$\textbf{98} & 92$\to$\textbf{96} & 94$\to$\textbf{98} \\
push buttons & 96$\to$\textbf{100} & 100$\to$74 & 90$\to$44 & 100$\to$68 & 98$\to$\textbf{100} & 100$\to$98 \\
put item in drawer & 0$\to$\textbf{6} & 0$\to$\textbf{4} & 0$\to$\textbf{10} & 6$\to$\textbf{8} & 0$\to$\textbf{2} & 0$\to$\textbf{0} \\
put knife on chopping board & 10$\to$\textbf{34} & 4$\to$\textbf{24} & 12$\to$10 & 8$\to$\textbf{32} & 20$\to$\textbf{20} & 22$\to$18 \\
put money in safe & 82$\to$\textbf{92} & 68$\to$\textbf{78} & 72$\to$\textbf{78} & 82$\to$\textbf{86} & 42$\to$\textbf{64} & 38$\to$\textbf{88} \\
put rubbish in bin & 50$\to$\textbf{62} & 8$\to$\textbf{66} & 50$\to$40 & 64$\to$52 & 22$\to$\textbf{30} & 36$\to$\textbf{36} \\
put umbrella in umbrella stand & 10$\to$\textbf{26} & 14$\to$\textbf{26} & 12$\to$\textbf{18} & 6$\to$\textbf{24} & 2$\to$\textbf{12} & 10$\to$\textbf{14} \\
reach and drag & 58$\to$\textbf{64} & 64$\to$\textbf{76} & 56$\to$\textbf{62} & 52$\to$\textbf{68} & 58$\to$\textbf{70} & 68$\to$\textbf{80} \\
reach target & 100$\to$\textbf{100} & 98$\to$\textbf{100} & 98$\to$\textbf{100} & 100$\to$\textbf{100} & 100$\to$\textbf{100} & 100$\to$\textbf{100} \\
slide block to target & 66$\to$\textbf{90} & 58$\to$\textbf{96} & 82$\to$\textbf{98} & 80$\to$\textbf{86} & 64$\to$\textbf{76} & 74$\to$\textbf{88} \\
stack blocks & 0$\to$\textbf{0} & 0$\to$\textbf{0} & 0$\to$\textbf{0} & 0$\to$\textbf{0} & 0$\to$\textbf{0} & 0$\to$\textbf{0} \\
take item out of drawer & 0$\to$\textbf{10} & 0$\to$\textbf{10} & 4$\to$\textbf{30} & 4$\to$\textbf{10} & 0$\to$\textbf{2} & 4$\to$0 \\
take lid off saucepan & 84$\to$\textbf{96} & 82$\to$\textbf{98} & 94$\to$\textbf{98} & 84$\to$\textbf{92} & 94$\to$80 & 90$\to$\textbf{96} \\
take umbrella out of umbrella stand & 92$\to$\textbf{98} & 88$\to$\textbf{98} & 86$\to$\textbf{100} & 90$\to$\textbf{94} & 92$\to$\textbf{96} & 100$\to$\textbf{100} \\
toilet seat down & 90$\to$\textbf{96} & 100$\to$98 & 96$\to$\textbf{100} & 92$\to$\textbf{98} & 92$\to$\textbf{94} & 90$\to$\textbf{96} \\
toilet seat up & 82$\to$\textbf{86} & 54$\to$\textbf{86} & 38$\to$\textbf{90} & 74$\to$\textbf{92} & 46$\to$\textbf{72} & 66$\to$\textbf{72} \\
turn oven on & 18$\to$\textbf{52} & 4$\to$\textbf{28} & 24$\to$\textbf{72} & 28$\to$\textbf{46} & 6$\to$\textbf{24} & 16$\to$\textbf{52} \\
turn tap & 54$\to$\textbf{70} & 54$\to$\textbf{64} & 78$\to$\textbf{82} & 74$\to$\textbf{84} & 26$\to$\textbf{54} & 56$\to$\textbf{68} \\
unplug charger & 24$\to$\textbf{30} & 6$\to$\textbf{30} & 40$\to$\textbf{66} & 30$\to$\textbf{70} & 14$\to$\textbf{14} & 12$\to$\textbf{12} \\
water plants & 70$\to$\textbf{74} & 38$\to$\textbf{56} & 50$\to$\textbf{86} & 64$\to$\textbf{80} & 42$\to$\textbf{56} & 58$\to$56 \\
\hline
\end{tabular}
\end{table}

\end{document}